\documentclass[10pt,twocolumn,letterpaper]{article}

\usepackage{iccv}
\usepackage{times}
\usepackage{graphicx}
\usepackage{amsmath}
\usepackage{amssymb}
\usepackage{multirow}
\usepackage{rotating}
\usepackage[pagebackref=true,breaklinks=true,letterpaper=true,colorlinks,bookmarks=false]{hyperref}
\usepackage{subcaption}
\usepackage{booktabs}
\usepackage{array}
\usepackage{flushend}
\usepackage{cite}
\usepackage{tikz}
\usepackage{xcolor}
\usepackage[toc,page]{appendix}

\iccvfinalcopy 

\def\iccvPaperID{668} 

\makeatletter
\renewcommand{\paragraph}{%
  \@startsection{paragraph}{4}%
  {\z@}{0.5ex \@plus 1ex \@minus .2ex}{-1em}%
  {\normalfont\normalsize\bfseries}%
}
\makeatother

\renewcommand{\Re}{\mathbb{R}}

\begin{document}


\title{Efficient Segmentation: \\ Learning Downsampling Near Semantic Boundaries}

\author{
Dmitrii Marin$^*$\hspace{10mm}
Zijian He$^\dag$\hspace{10mm}
Peter Vajda$^\dag$\hspace{10mm}
Priyam Chatterjee$^\dag$\hspace{10mm}\\
Sam Tsai$^\dag$\hspace{10mm}
Fei Yang$^\ddag$\hspace{10mm}
Yuri Boykov$^*$\hspace{10mm}
\\
$^*$University of Waterloo, Canada
\hspace{15mm} 
$^\dag$Facebook Inc., USA
\hspace{15mm}
$^\ddag$TAL Education, China
\\ 
{\tt\footnotesize \{d2marin,yboykov\}@uwaterloo.ca  \hspace{5mm}  \{zijian,vajdap,priyamc,sstsai\}@fb.com \hspace{5mm} yang.fei@100tal.com}
}

\maketitle
\thispagestyle{empty}

\begin{abstract}
   Many automated processes such as auto-piloting rely on a good semantic segmentation as a critical component. To speed up performance, it is common to downsample the input frame. However, 
   this comes at the cost of missed small objects and reduced accuracy at semantic boundaries. To address this problem, we propose a new content-adaptive downsampling technique that learns to favor sampling locations near semantic boundaries of target classes. Cost-performance analysis shows that our method consistently outperforms the uniform sampling improving balance between accuracy and computational efficiency. 
   Our adaptive sampling gives segmentation with better quality of boundaries and more reliable support for smaller-size objects.
\end{abstract}

\section{Introduction}

Recent progress in hardware technology has made running efficient deep learning models on mobile devices possible. This has enabled many on-device experiences relying on deep learning-based computer vision systems. However, many tasks including semantic segmentation still require downsampling of the input image trading off accuracy in finer details for better inference speed~\cite{howard2017mobilenets,zhao2017icnet}. We show that uniform downsampling is sub-optimal and propose an alternative content-aware adaptive downsampling technique driven by semantic boundaries. We hypothesize that for better segmentation quality more pixels should be picked near semantic boundaries. With this intuition, we formulate a neural network model for learning content-adaptive sampling from ground truth semantic boundaries, see Fig.~\ref{fig:nus}.

\newcommand{\zoom}[1]{
    \begin{tikzpicture}
        \node[inner sep=0pt] (russell) at (0,0) 
            {\includegraphics[width=0.48\linewidth]{#1}};
        \node[above right,inner sep=0pt] (zoom) at (0.04\linewidth,0.04\linewidth) 
            {\fcolorbox{green}{white}{\fcolorbox{green}{white}{\fcolorbox{green}{white}{\includegraphics[width=0.2\linewidth,clip=true,trim=4.1cm 3.0cm 7.3cm 8.38cm]{#1}}}}};
        \draw [->,>=stealth,thick,color=green] (zoom) -- (-1.75ex,-3.75ex) node (A) {};
        \draw [color=green] (A) rectangle +(-3ex,-3ex);
    \end{tikzpicture} 
}

\begin{figure}[t]
    \centering
    \setlength\tabcolsep{0pt}
    \setlength\fboxsep{0pt}
    \begin{tabular}{cc}
    \zoom{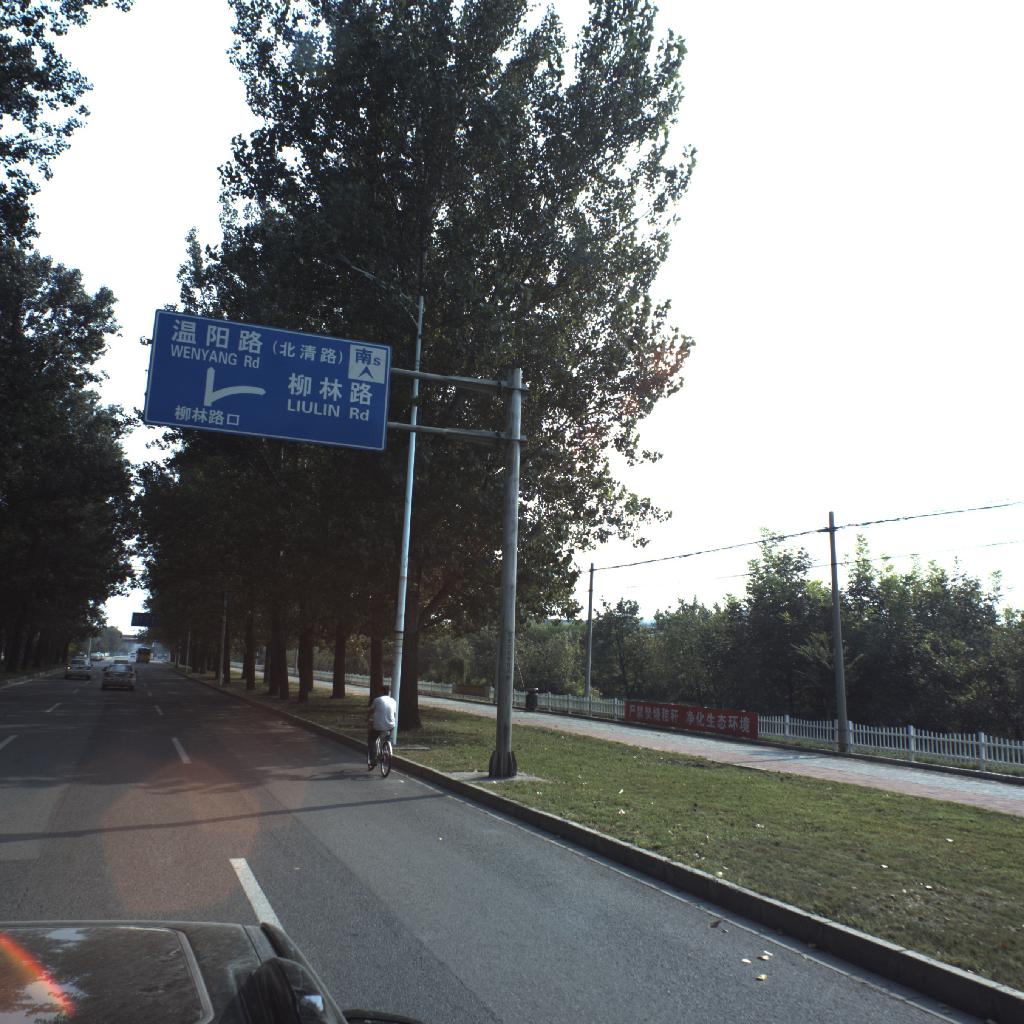}
    &
    \zoom{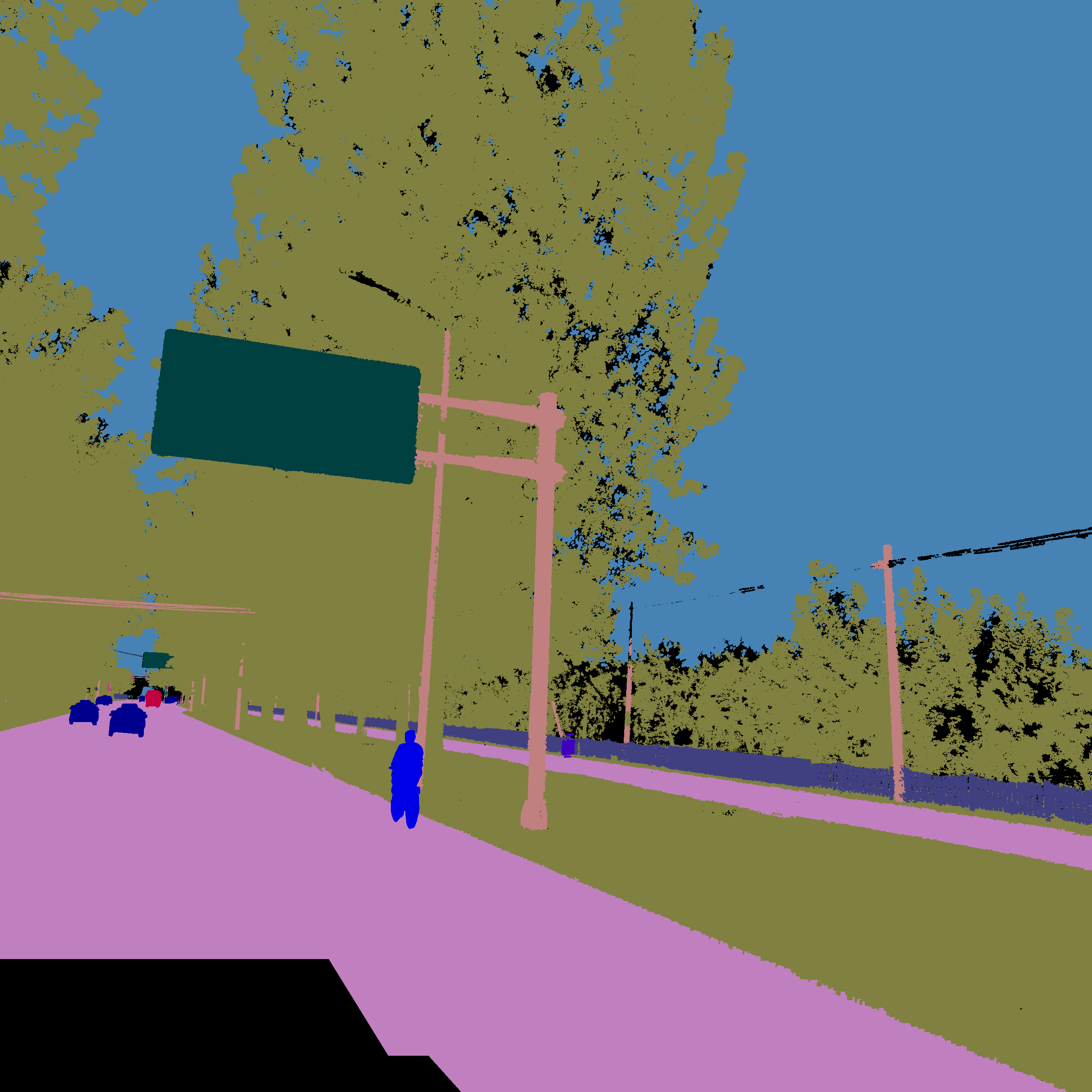}
     \\[-1ex]
     {\small \centering (a) original $2710\!\times\!2710$ image} & {\small (b) ground truth labels} 
     \\[1ex]
    \zoom{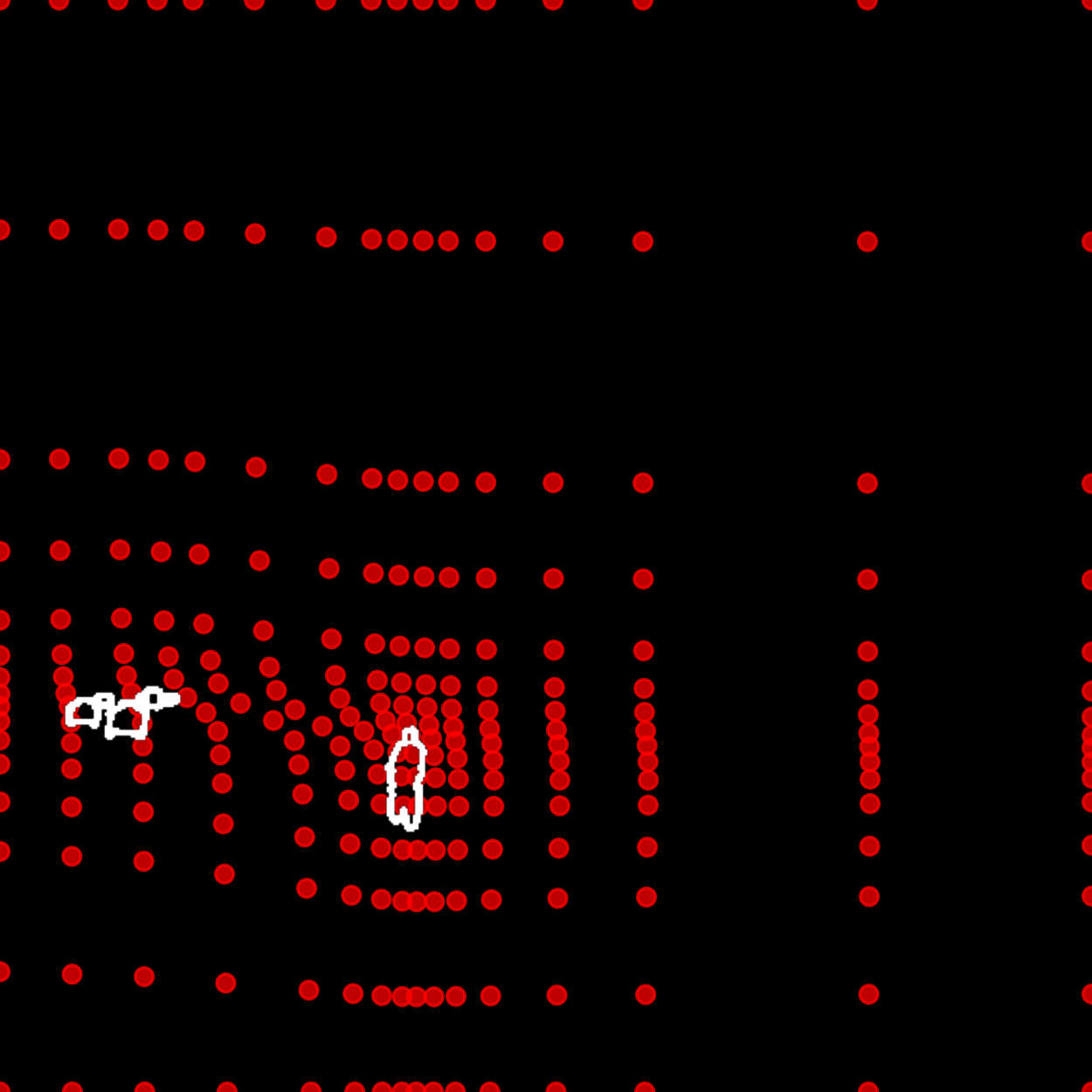}
    &
    \zoom{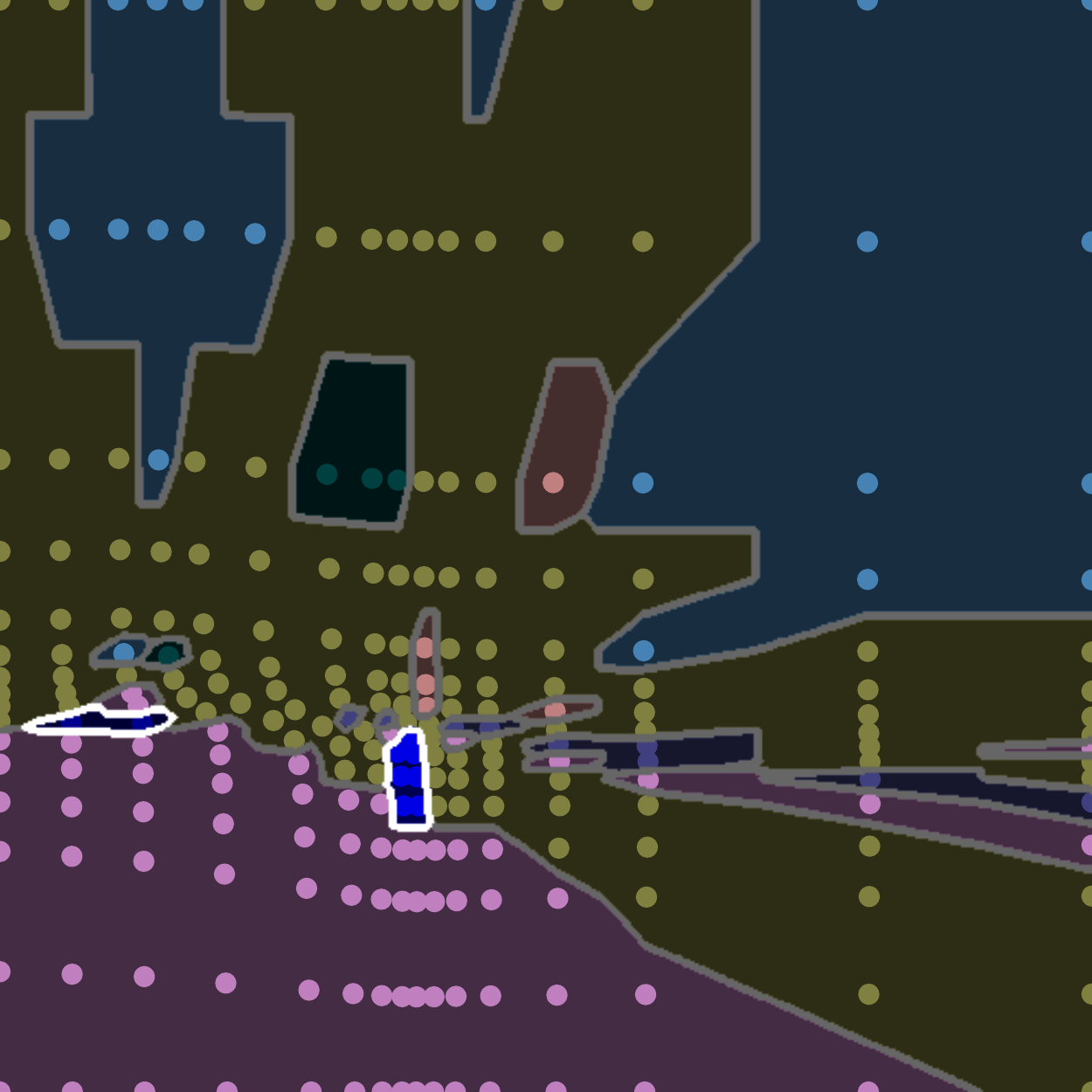} 
     \\[-0.7ex]
    \parbox{27ex}{\centering  \small (c) target semantic boundaries \\  and adaptive sampling locations} & \parbox{27ex}{\small  \centering (d)  interpolation of sparse classifications (target is white)}
    \end{tabular}
    \caption{Illustration of our content-adaptive downsampling method on a high-resolution image (a). Given ground truth (b), we compute a non-uniform grid of sampling locations (red in (c))
    pulled towards semantic boundaries of target classes (white in (c)). We use these points for training our {\em auxiliary network} to automatically produce such sparsely sampled
    locations. Their classification (colored dots in (d)) can be produced by a separately trained efficient low-res segmentation CNN.  
    Concentration of sparse classifications near boundaries of target classes improves accuracy of interpolation~(d) compared to uniform sampling, see Fig.~\ref{fig:qualitative examples}. 
    }
    \vspace{-3ex} 
    \label{fig:nus}
\end{figure}

The advantages of our non-uniform downsampling over the uniform one are two-fold. First, the common uniform downsampling complicates accurate localization of boundaries in the original image. Indeed, assuming $N$ uniformly sampled points over an image of diameter $D$, the distance between neighboring points
gives a bound for the segmentation boundary localization errors $\mathcal O(\frac{D}{\sqrt{N}})$. 
In contrast, analysis in \ref{sec:supp} shows that the error bound decreases significantly faster with respect to the number of sample points 
$\mathcal O(\frac{\kappa l^2}{N^2})$ assuming they are uniformly distributed near the segment boundary of max curvature $\kappa$ and length~$l$.
Our non-uniform boundary-aware sampling approach selects more pixels around semantic boundaries reducing quantization errors on the boundaries. 

Second, our non-uniform sampling implicitly accounts for scale variation via reducing the portion of the downsampled image occupied by larger segments and increasing that of smaller segments. It is well-known that presence of the same object class at different scales complicates automatic image understanding \cite{he2004multiscale, russell2009associative, Girshick_2014_CVPR, Girshick_2015_ICCV, yu2015multi, chen2018deeplab, chen2017rethinking, zhao2017pyramid, chen2018encoder}. Thus, the scale equalizing effect of our adaptive downsampling simplifies learning. As shown in Fig.~\ref{fig:nus}(c,d), our approach samples many pixels inside the cyclist, while the uniform downsampling may miss that person all together. 

\begin{figure}[!t]
   \centering
   \includegraphics[width=\linewidth]{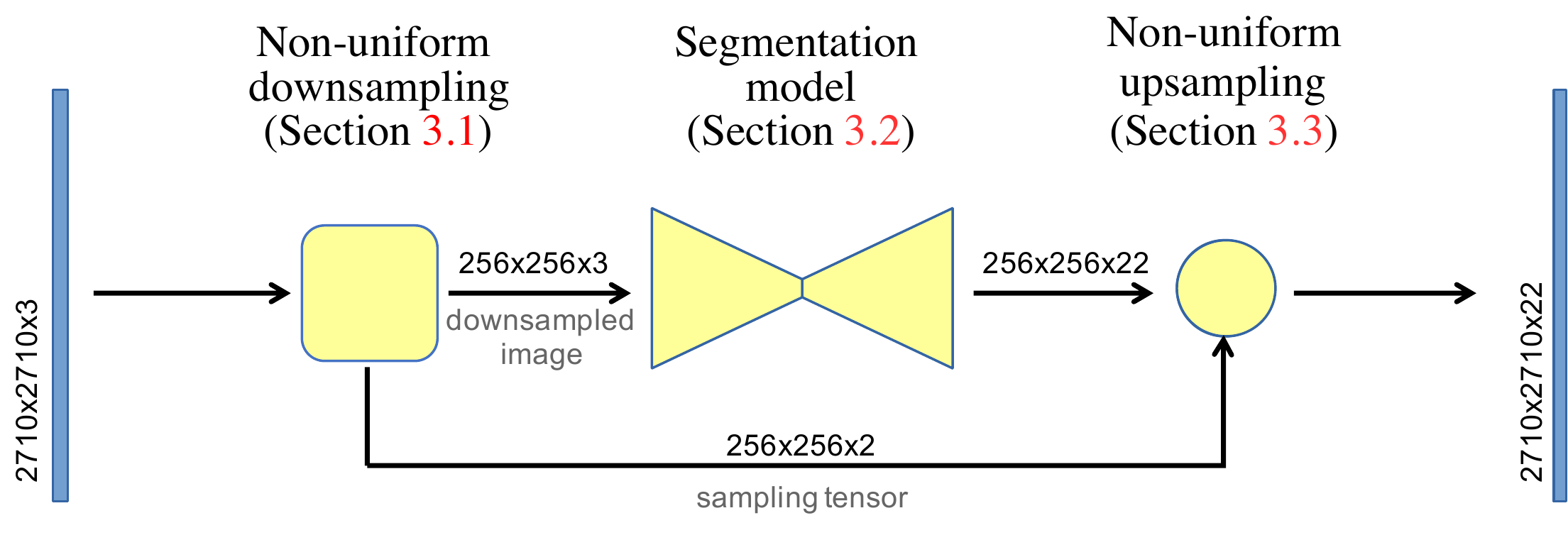}
   \caption{Proposed efficient segmentation architecture with adaptive downsampling. The first block (detailed in Fig.\ref{fig:architecture_sampling})
   takes a high-res image and outputs sampling locations (\emph{sampling tensor}) and a downsampled image. 
   Then, the downsampled image is segmented by some standard model. 
   Finally, the result is upsampled to the original resolution. }
\label{fig:architecture}
\end{figure}

With the proposed content-adaptive sampling, a semantic segmentation system consists of three parts, see Fig.\ref{fig:architecture}. The first is our non-uniform downsampling block trained to sample pixels near semantic boundaries of target classes. The second part segments the downsampled image and can be based on practically any existing segmentation model. The last part upsamples the segmentation result producing a segmentation map at the original (or any given) resolution. Since we need to invert the non-uniform sampling, standard CNN interpolation techniques are not applicable. 

Our contributions in this paper are as follows: 
\begin{itemize}
    \item We propose adaptive downsampling aiming at accurate representation of targeted semantic boundaries. 
    We use an efficient CNN to reproduce such downsampling.
    \item Most segmentation architectures can benefit from non-uniform downsampling by incorporating our content-adaptive sampling and interpolation components.
    \item We apply our framework to semantic segmentation and show consistent improvements on many architectures and datasets. Our cost-performance analysis accounts for the computational overhead. We also analyze improvements from our adaptive downsampling at semantic boundaries and on objects of different sizes.
\end{itemize}

Sec.~\ref{sec:prior_work} provides an overview of prior works. Sec.~\ref{sec:nus} describes our approach in details. Sec.~\ref{sec:exp} compares many state-of-the-art semantic segmentation architectures with uniform and our adaptive downsampling on multiple datasets.

\section{Prior work}
\label{sec:prior_work}

Semantic segmentation requires a class assignment for each pixel in an image. This problem is important for many automated navigational applications. We first review some related literature on this topic. We then provide a brief review of some relevant non-uniform sampling methods.


Many segmentation networks are built upon basic image classification networks, \eg \cite{long2015fully,wu2016wider,chen2018deeplab,chen2017rethinking,zhao2017pyramid,chen2018encoder}. These approaches modify the base model to produce dense higher resolution features maps. For example, Long \etal \cite{long2015fully} used \emph{fully convolutional network} \cite{lecun1995convolutional} and trainable deconvolution layers producing higher resolution dense feature maps. They also note that \emph{algorithme \`a trous}, a technique well knowing in signal processing~\cite{holschneider1990real}, is a way to increase resolution of the feature maps. This idea was studied in \cite{chen2018deeplab} where the authors introduced \emph{dilated convolutions} that allow removal of max pooling layers from a trained model producing higher resolution feature maps with higher field of view without the need to retrain the model.

\begin{table}[!b]
    \centering
    \footnotesize
    \begin{tabular}{m{12ex}|c|c|c}
        & \textbf{two-stage} & \textbf{ours} & \textbf{single-stage} \\
        & \footnotesize {\cite{Girshick_2014_CVPR,Girshick_2015_ICCV,he2017mask}} & Sec.~\ref{sec:nus} & \footnotesize {\cite{long2015fully,ronneberger2015u,chen2018deeplab}} \\
        \hline
        \bf accuracy & ++ & + & - \\
        \hline
        \bf speed & - & + & ++ \\
        \hline
        \bf multi-object speed & - - & + & ++ \\
        \hline
        \bf simplicity & - & + & ++ \\
        \hline
        \bf multi-scale & ++ & + & - \\
        \hline
        \bf boundary precision & ++ & + & -
    \end{tabular}
    \caption{Segmentation approaches with pros \& cons (+/-).}
    \label{tab:approach_comparison}
\end{table}

Segmentation models built upon classification models inherit one limiting property, that is the base classification models \cite{krizhevsky2012imagenet,simonyan2014very,he2016deep} tend to have many features in the deeper layers. That results in an extensive resources consumption when increasing the resolution of later feature maps (using for example \emph{algorithme \`a trous} \cite{holschneider1990real}). As a result, the final output is typically chosen to be of lower resolution, with an interpolation employed to upscale the final score map. 

The alternative direction for segmentation (and more generally for pixel-level prediction) is based on ``hourglass models'' that first produce low resolution deep features and then gradually upsample the features employing common network operations and skip connections \cite{ronneberger2015u, badrinarayanan2015segnet,noh2015learning,newell2016stacked}.

The need and advantage of aggregating information from different scales have been long recognized in the computer vision literature \cite{he2004multiscale,russell2009associative,Girshick_2014_CVPR,Girshick_2015_ICCV,yu2015multi,xia2016zoom,chen2018deeplab,chen2017rethinking,zhao2017pyramid,chen2018encoder}. 
One way to tackle the multiscale challenge is to first detect the location of objects and then segment the image using either a cropped original image \cite{Girshick_2014_CVPR,xia2016zoom} or cropped feature maps \cite{Girshick_2015_ICCV,he2017mask}. These two-stage approaches separate the problems of scale learning and segmentation, making the latter task easier. As a result the accuracy of segmentation improves and instance level segmentation is straightforward. However, such an approach comes with a significant computational cost when many objects are present since each object needs to be segmented individually. Our method improves upon the single-stage approach for a small computational cost and, thus, is positioned in between of these two approaches. Table \ref{tab:approach_comparison} outlines pros and cons of our approach compared with two-stage and single-stage methods.


Spatial Transformer Networks \cite{jaderberg2015spatial,recasens2018learning} learn spatial transformations (warping) of the CNN input. They explore different parameterizations for spatial transformation including affine, projective, splines \cite{jaderberg2015spatial} or specially designed saliency-based layers  \cite{recasens2018learning}. Their focus is to undo different data distortions or to ``zoom-in'' on salient regions, while our approach is focused on efficient downsampling retaining as much information around semantic boundaries as possible. They do not use their approach in the context of pixel-level predictions (\eg segmentation) and do not consider the inverse transformations (Sec.~\ref{sec:nus:upsample} in our case).

Deformable convolutions \cite{dai2017deformable,jeon2017active} augment the spatial sampling locations in the standard convolutions with additional adaptive offsets. In their experiments the deformable convolutions replace traditional convolutions in the last few layers of the network making their approach complementary to ours. The goal is to allow the new convolution to pick the features from the best locations in the previous layer. Our approach focuses on choosing the best locations in the original image and thus has access to more information.

Other complementary approaches include skipping some layers at some pixels \cite{figurnov2016perforatedcnns} and early stopping of network computation for some spatial regions of the image \cite{figurnov2017spatially,li2017not}. Similarly, these methods modify computation at deeper network layers and do not concern image downsampling.


Pascal \etal~\cite{Weickert2016compression} showed that an advanced extrapolation method (based on PDEs) applied to a smartly selected small number of pixels reproduces the original image with low error leading to a state-of-the-art compression scheme. Their method selects pixels around strong edges of the image. In contrast, we do not use edges of the image when deciding sampling locations. Instead, we rely on machine learning based on semantic boundaries to predict sampling locations.

Adaptive sampling is also employed in curve and surface approximations and splines reduction~\cite{TILLER1992445,obeidat2009intelligent,HERNANDEZMEDEROS2003363}.

\section{Boundary Driven Adaptive Downsampling}\label{sec:nus}

Fig.~\ref{fig:architecture} shows three main stages of our system: content-adaptive downsampling, segmentation and upsampling. The downsampler, described in Sec.~\ref{sec:nus:sampling_model}, determines non-uniform sampling locations and produces a downsampled image. The segmentation model then processes this (non-uniformly) downsampled image. We can use any existing segmentation model for this purpose. The results are treated as sparsely classified locations in the original image. The third part, described in Sec.~\ref{sec:nus:upsample}, uses interpolation to recover segmentation at the original resolution, see Fig.~\ref{fig:nus}(d). 

Let us introduce notation. Consider a high-resolution image ${I}=\{I_{ij}\}$ of size $H \times W$ with $C$ channels. Assuming relative coordinate system, all pixels have spatial coordinates that form a uniform grid covering square $[0,1]^2$. Let $I[u,v]$ be the value of the pixel that has spatial coordinates closest to $(u,v)$ for $u,v\in[0,1]$. Consider tensor ${ \phi} \in [0,1]^{2\times h\times w}$. We denote elements of ${ \phi}$ by $\phi^c_{ij}$ for $c\in\{0,1\}$, $i \in \{1,2,\dots,h\}$, $j \in \{1,2,\dots,w\}$. We refer to such tensors as \emph{sampling tensors}. Let ${ \phi}_{ij}$ be the point $(\phi^0_{ij},\phi^1_{ij})$. Fig.~\ref{fig:nus}(c) shows an example of such points.

The sampling operator
\begin{equation*}
\;\; \Re^{C\times H \times W} \times [0,1]^{2 \times h \times w} \;\; \to \;\; \Re^{C\times h \times w}
\end{equation*}
maps a pair of image ${I}$ and sampling tensor ${ \phi}$ to the corresponding sampled image $J=\{ J_{ij} \}$ such that 
\begin{equation}\label{eq:sampling operator} 
J_{ij}:= I[\phi^0_{ij}, \phi^1_{ij}].
\end{equation} 

The uniform downsampling can be defined by a sampling tensor $u \in [0,1]^{2\times h\times w}$ such that $u^0_{ij}=(i-1)/(h-1)$ and $u^1_{ij}=(j-1)/(w-1)$. 

\subsection{Sampling Model} \label{sec:nus:sampling_model}

Our non-uniform sampling model should balance between two competing objectives. On one hand we want our model to produce finer sampling in the vicinity of semantic boundaries. On the other hand, the distortions due to the non-uniformity should not preclude successful segmentation of the non-uniformly downsampled image.

\begin{figure}
    \centering
    \setlength\tabcolsep{1pt}
    \setlength\fboxsep{0pt}
    \begin{tabular}{ccccc}
        \includegraphics[width=0.245\linewidth]{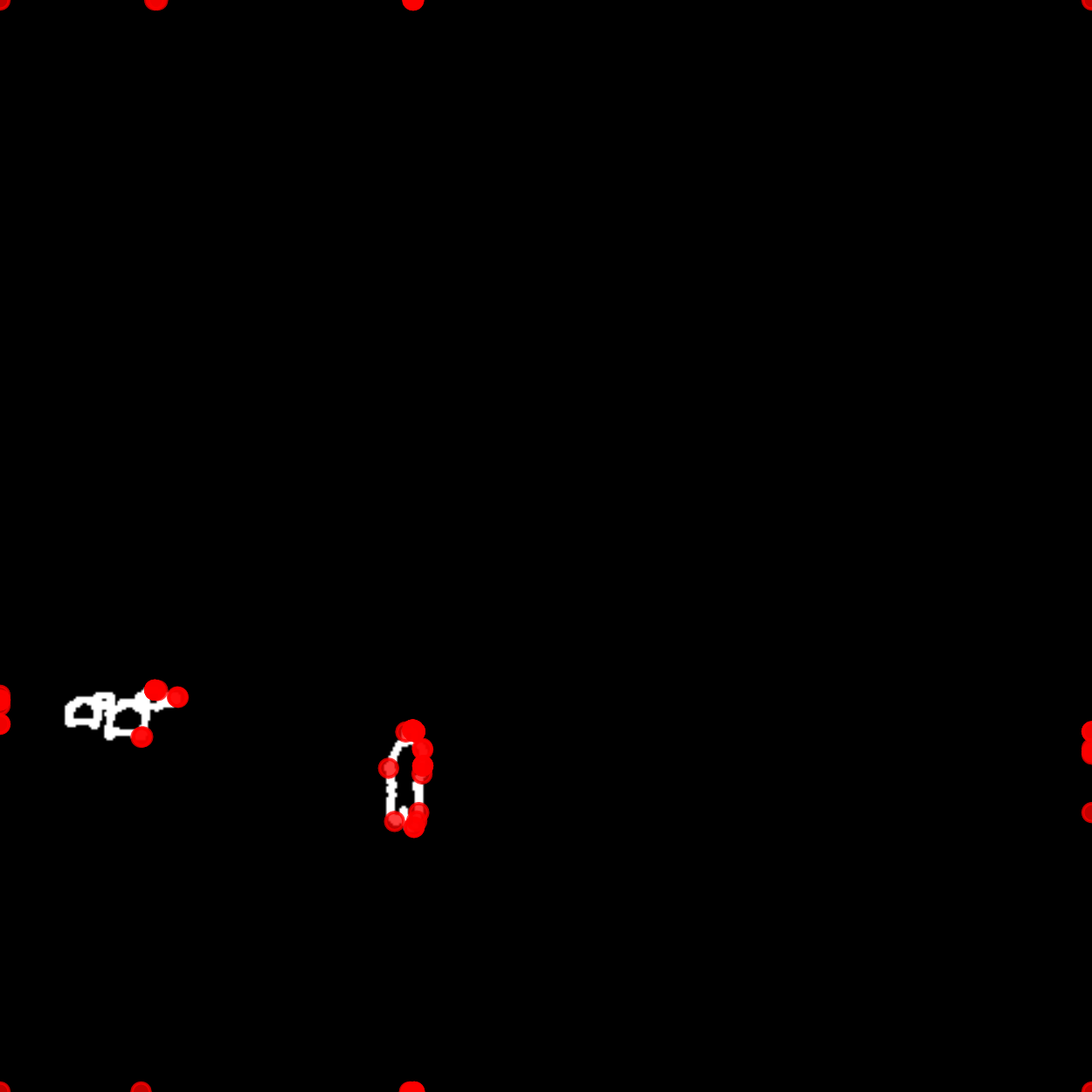}
        &
        \includegraphics[width=0.245\linewidth]{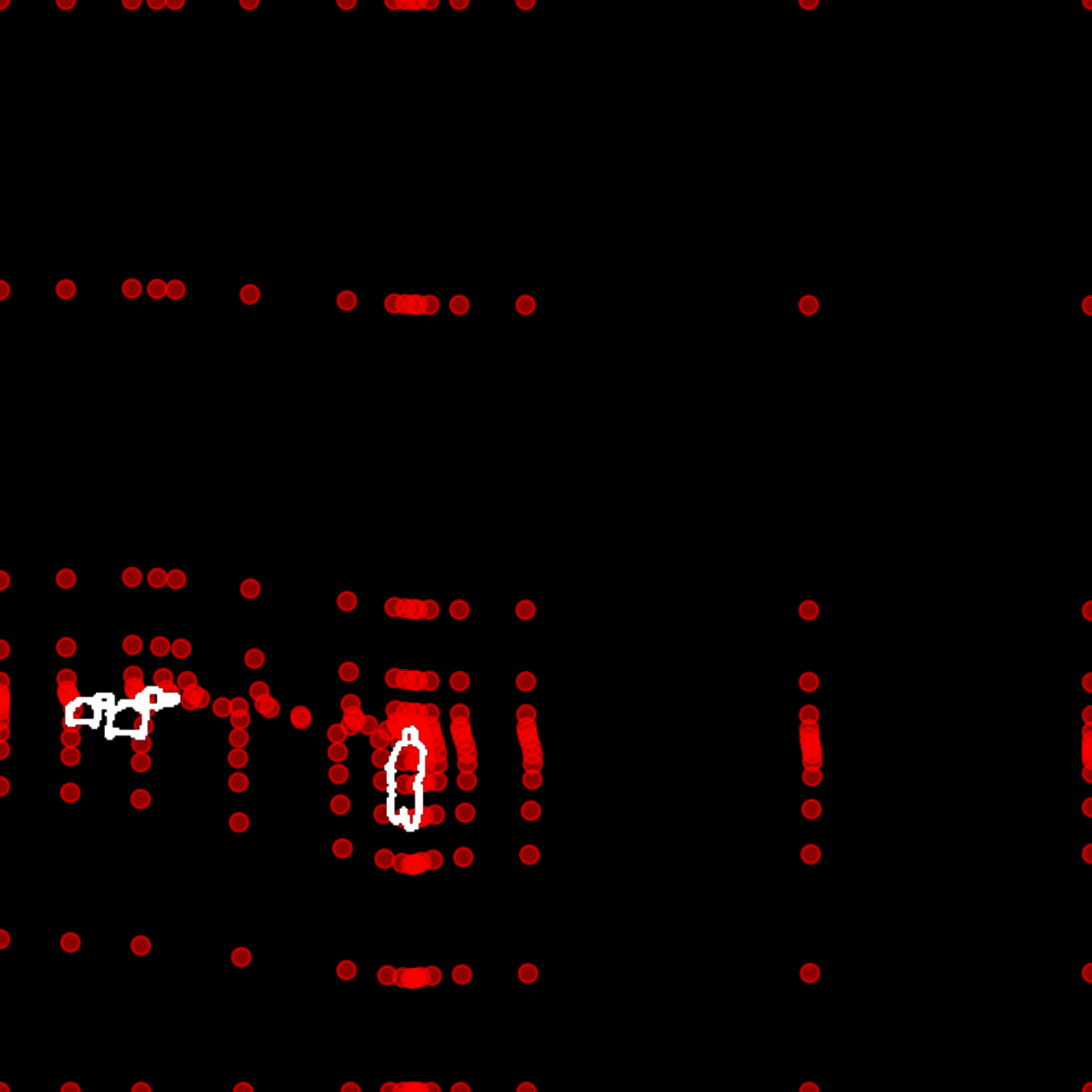}
        &
        \includegraphics[width=0.245\linewidth]{sampling_examples/001167_anchors_boundaries_2.pdf}
        &
        \includegraphics[width=0.245\linewidth]{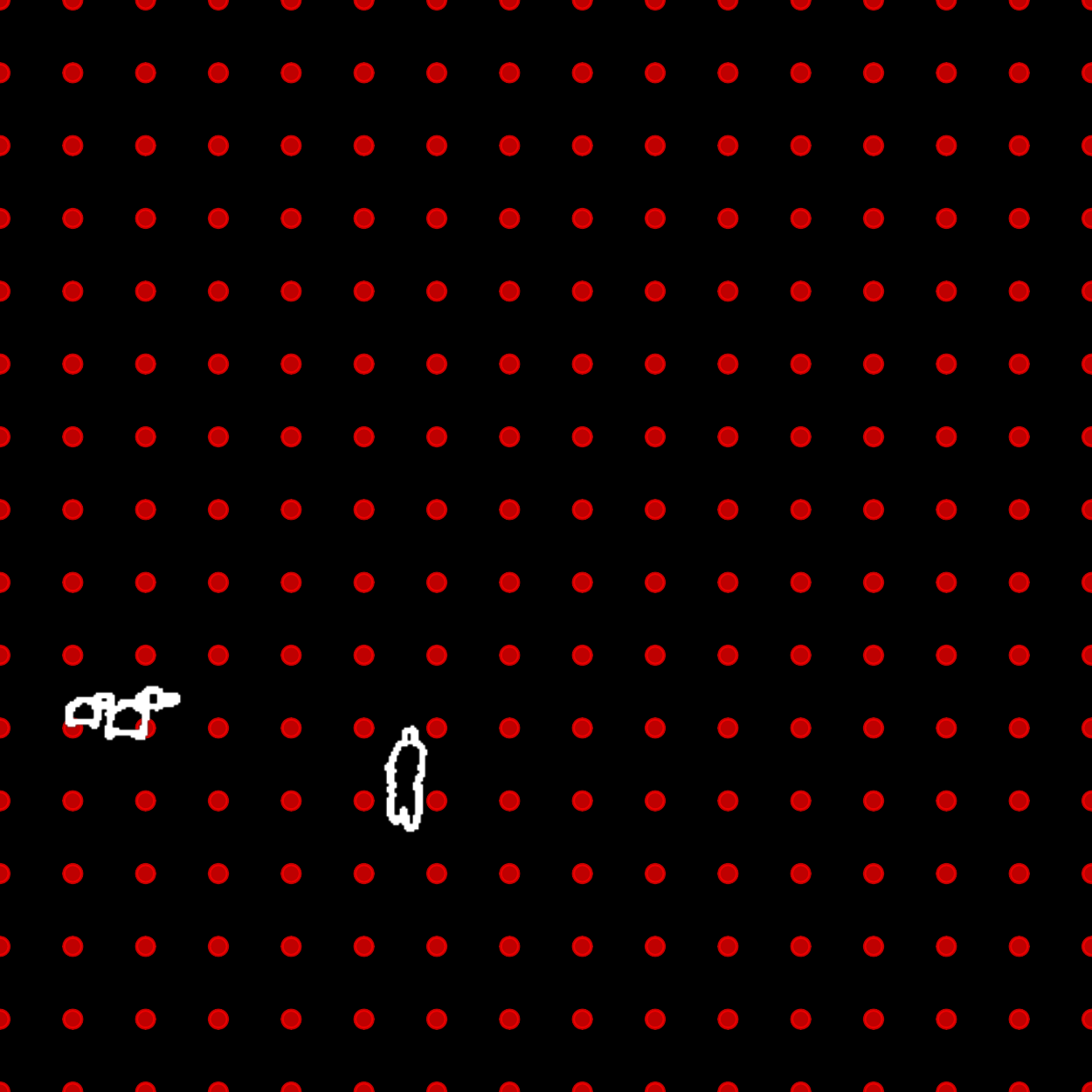}
        \\[-0.24\linewidth]
          \large $\color{yellow}\boldsymbol{\lambda=0}$ 
        & \large $\color{yellow}\boldsymbol{\lambda=0.5}$ 
        & \large $\color{yellow}\boldsymbol{\lambda=1}$ 
        & \large $\color{yellow}\boldsymbol{\lambda=+\infty}$
        \\[0.15\linewidth]
    \end{tabular}
    \caption{Boundary driven sampling for different $\lambda$ in~\eqref{eq:energy}. Extreme $\lambda$ sample either semantic boundaries (left) or uniformly (right). Middle-range $\lambda$ yield in-between sampling.}
    \label{fig:lambda}
\end{figure}

Assume for image ${I}$ (Fig.~\ref{fig:nus}(a)) we have the ground truth semantic labels (Fig.~\ref{fig:nus}(b)). We compute a boundary map (white in Fig.~\ref{fig:nus}(c)) from the semantic labels. Then for each pixel we compute the closest pixel on the boundary. Let $b(u_{ij})$ be the spatial coordinates of a pixel on the semantic boundary that is the closest to coordinates $u_{ij}$ (distance transform). We define our content-adaptive non-uniform downsampling as sampling tensor $\phi$ minimizing the energy
\begin{equation}\label{eq:energy}
    E({\bf \phi}) = \sum_{i,j}\left\| \phi_{ij} - b(u_{ij}) \right\|^2 + \lambda \!\!\!\! \sum_{\substack{|i-i'|+\\|j-j'|=1}}  \left\| \phi_{ij} - \phi_{i'j'} \right\|^2
\end{equation}
subject to \emph{covering constraints}
\begin{equation}\label{eq:constraints}
\begin{array}{ll}
    \phi \in [0,1]^{2\times h \times w} \\
    \phi^0_{1\,j} = 0 \;\;\;\&\;\;\; \phi^0_{hj} = 1, & 1 \le j \le w, \\
    \phi^1_{i\,1} = 0 \;\;\;\&\;\;\; \phi^1_{iw} = 1, & 1 \le i \le h. \\
\end{array}
\end{equation}
The first term in~\eqref{eq:energy} ensures that sampling locations are close to semantic boundaries, while the second term ensures that the spatial structure of the sampling locations is not distorted excessively. The constraints provide that the sampling locations cover the entire image. 
This least squares problem with convex constraints can be efficiently solved globally via a set of sparse linear equations. Red dots in Figs.~\ref{fig:nus}(c) and~\ref{fig:lambda} illustrate solutions for different values of $\lambda$.

We train a relatively small auxiliary network to predict the sampling tensor without boundaries. The auxiliary network can be significantly smaller than the base segmentation model as it solves a simpler problem. It learns cues indicating presence of the semantic boundaries. For example, the vicinity of vanishing points is more likely to contain many small objects (and their boundaries). Also, small mistakes in the sampling locations are not critical as the final classification decision is left for the segmentation network.

\begin{figure*}[t]
\begin{center}
   \includegraphics[width=\linewidth,trim=0 2mm 0 0]{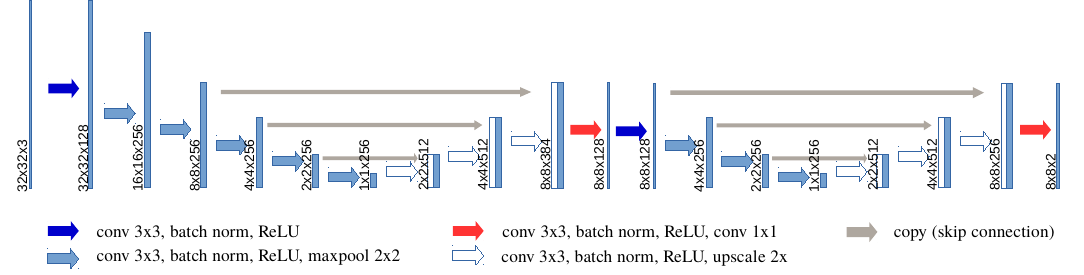}
   \vspace{-5mm}
\end{center}
   \caption{Double U-Net model for predicting sampling parameters. The depth of the first sub-network can vary (depending on the input resolution). The structure of the second sub-network is kept fixed. To improve efficiency, we use only one convolution (instead of two in~\cite{ronneberger2015u}) in each block. The number of features is $256$ in all layers except the first and the last one. We also use padded convolutions to avoid shrinking of feature maps, and we add batch normalization after each convolution.}
\label{fig:predict_sampling}
\end{figure*}

As an auxiliary network, we propose two U-Net~\cite{ronneberger2015u} sub-networks stacked together (Fig.~\ref{fig:predict_sampling}). The motivation for stacking sub-networks is to model the sequential processes of boundary computation and sampling points selection. We train this network with squared L2 loss between the network prediction and a tensor ``proposal'' $\tilde\phi=\arg\min_\phi E(\phi)$ minimizing \eqref{eq:energy} subject to \eqref{eq:constraints}\footnote{The network prediction is projected onto constraints \eqref{eq:constraints} during testing.
}.  Alternatively, one can directly use objective \eqref{eq:energy} as a regularized loss function \cite{weston2012deep,tang2018regularized}. Our proposal generation approach can be seen as a one step of ADM procedure for such a loss~\cite{marin2019beyond}. 

Once the sampling tensor is computed the original image is downsampled via sampling operator~\eqref{eq:sampling operator}.
Application of sampling tensor $\phi$ of size $(2,h,w)$ yields sampled image of size $h \times w$. If this is not the desired size $h'\times w'$ of downsampled image, we still can employ $\phi$ for sampling. To that end, we obtain a new sampling tensor $\phi'$ of shape $(2,h',w')$ by resizing $\phi$ using bilinear interpolation, see example in Fig.\ref{fig:tensor resizing}. 

\begin{figure}
   \centering
   \includegraphics[width=\linewidth,trim=0 6mm 0 17mm]{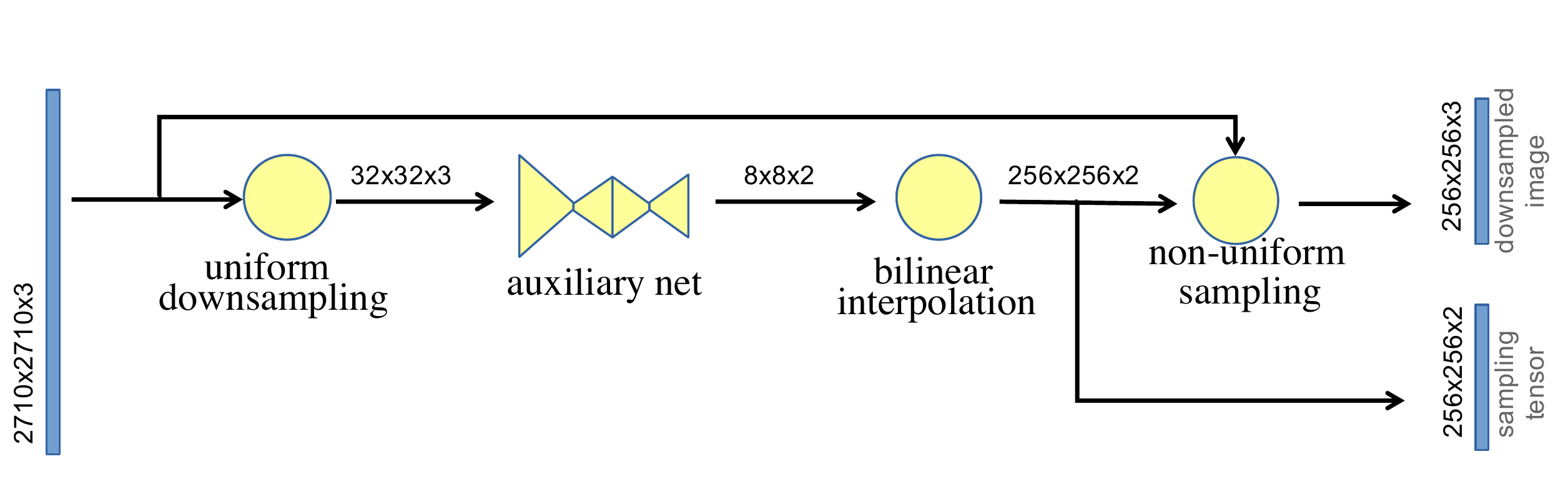}
   \caption{Architecture of non-uniform downsampling block in Fig.~\ref{fig:architecture}. A high-resolution image (\eg $2710 \times 2710$) is uniformly downsampled to a small image (\eg $ 32 \times 32 $) and then processed by an auxiliary network producing sampling locations stored in a \emph{sampling tensor}. This tensor is resized (Fig.~\ref{fig:tensor resizing}) and then used for non-uniform downsampling. }
   \label{fig:architecture_sampling}
\end{figure}

\begin{figure}
    \centering
    \setlength{\tabcolsep}{3pt}
    \renewcommand{\arraystretch}{0}
    \begin{tabular}{m{0.65\linewidth}m{0.3\linewidth}}
    \caption{An example of $8\times8$ sampling locations (red crosses) produced by an auxiliary network  and the result of resizing the corresponding \textit{sampling tensor} by the factor of $2$ (blue points) via bilinear interpolation, see Fig.~\ref{fig:architecture_sampling}.}
    \label{fig:tensor resizing} &
    \includegraphics[width=\linewidth,trim=0 0 0 15mm]{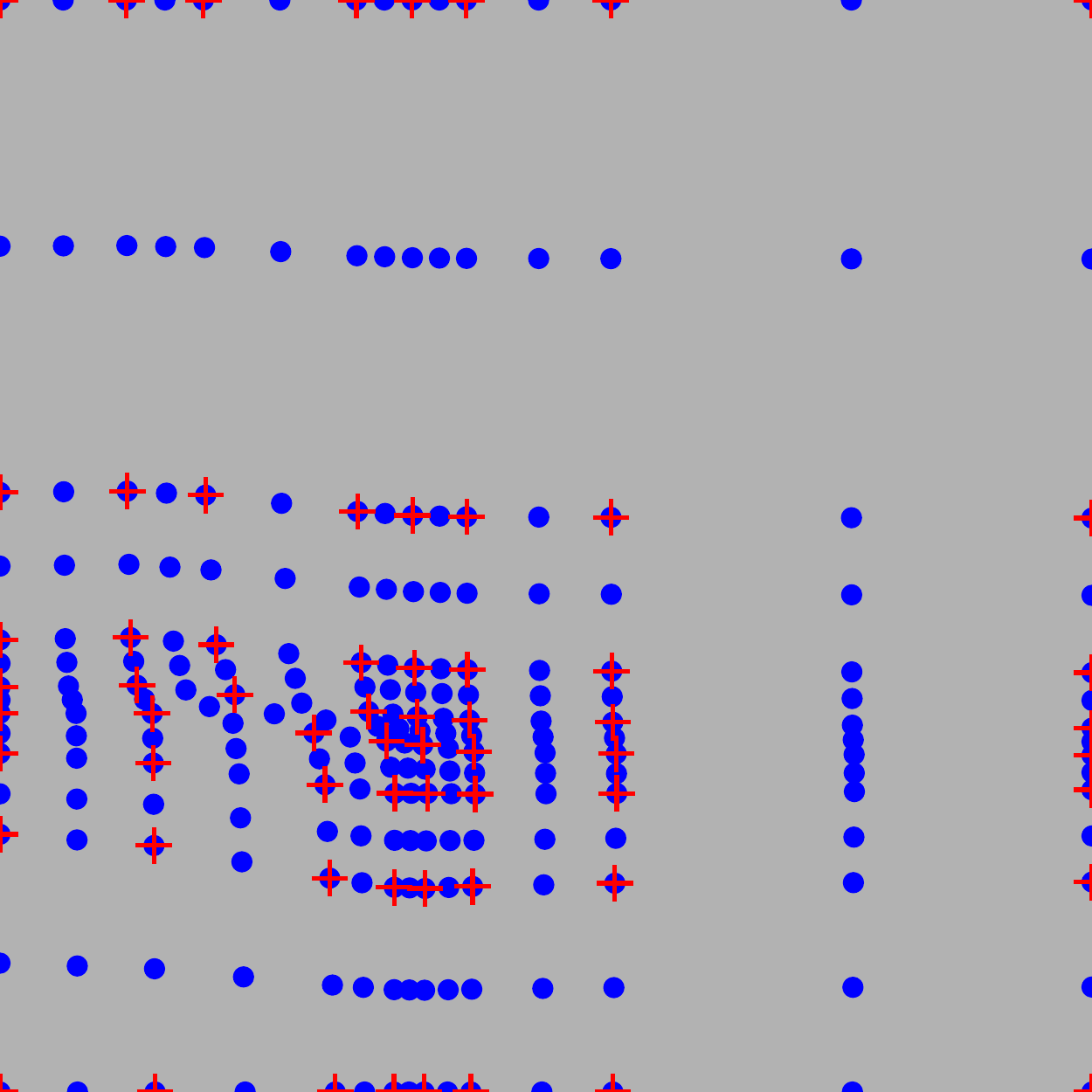}
    \end{tabular}
\end{figure}

Fig.\ref{fig:architecture_sampling} shows the architecture of our downsampling block.

\subsection{Segmentation Model}\label{sec:nus:base_model}
Our adaptive downsampling can be used with any off-the-shelf segmentation model as it does not place any constraints on the base segmentation model. Our improved results with base multiple models (U-Net~\cite{ronneberger2015u}, \mbox{PSP-Net}~\cite{zhao2017pyramid} and Deeplabv3+~\cite{chen2018encoder}) in Sec.~\ref{sec:exp} showcase this versatility.


\subsection{Upsampling}\label{sec:nus:upsample}


In keeping with prior work, we assume that the base segmentation model produces a final score map of the same size as its downsampled input. Thus, we need to upsample the output to match the original input resolution. In case of standard downsampling this step is a simple upscaling, commonly performed via bilinear interpolation. In our case, we need to ``invert'' the non-uniform transformation. Covering constraints \eqref{eq:constraints} ensure that the convex hall of the sampling locations covers the entire image, thus we can use interpolation to recover the score map at the original resolution. We use Scipy~\cite{scipygriddata} to interpolate the unstructured multi-dimensional data, which employs Delaunay~\cite{delaunay1934sphere} triangulation and barycentric interpolation within triangles \cite{Shirley2005fundamentals}.

An important aspect of our content-adaptive downsampling method in Sec.~\ref{sec:nus:sampling_model} is that it preserves the grid topology. Thus, an efficient implementation can skip the triangulation step and use the original grid structure. The interpolation problem reduces to a computer graphics problem of rendering a filled triangle, which can be efficiently solved by Bresenham's algorithm~\cite{Shirley2005fundamentals}.

\section{Experiments}
\label{sec:exp}

\begin{figure}
    \centering
    \footnotesize 
    \setlength{\tabcolsep}{0pt}
    \renewcommand{\arraystretch}{0.1}
    \begin{tabular}{m{0.25\linewidth}m{0.25\linewidth}m{0.25\linewidth}m{0.25\linewidth}}
        \parbox{\linewidth}{\centering (a) image \& our \\[-0.5ex] sampling \\[-0.5ex] locations} & 
        \parbox{\linewidth}{\centering (b) ground truth} & 
        \parbox{\linewidth}{\centering (c) predictions \\[-0.5ex]with  uniform \\[-0.5ex] downsampling} & 
        \parbox{\linewidth}{\centering (d) predictions \\[-0.5ex]with  our adaptive \\[-0.5ex] downsampling} \\
        \includegraphics[width=\linewidth]{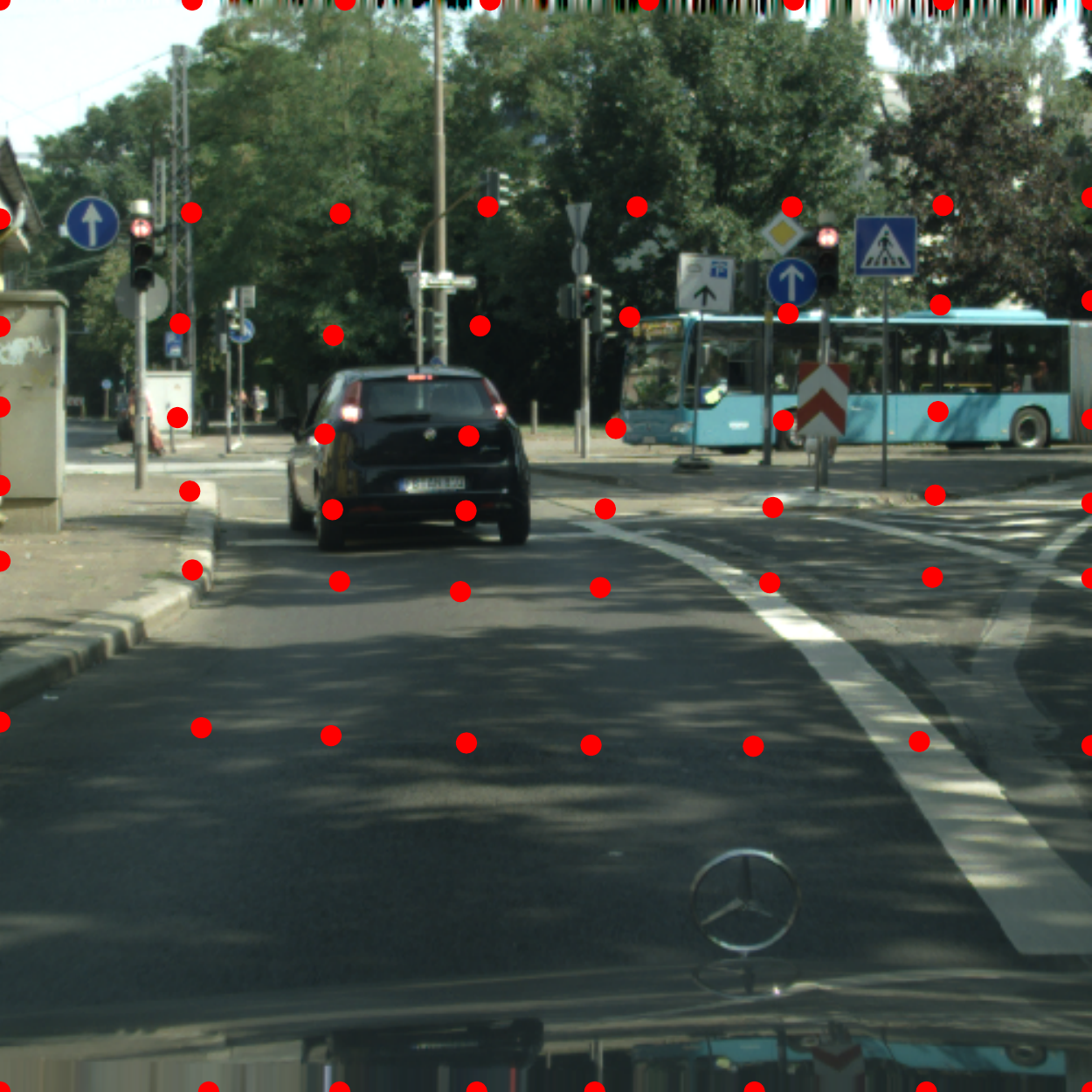} &
        \includegraphics[width=\linewidth]{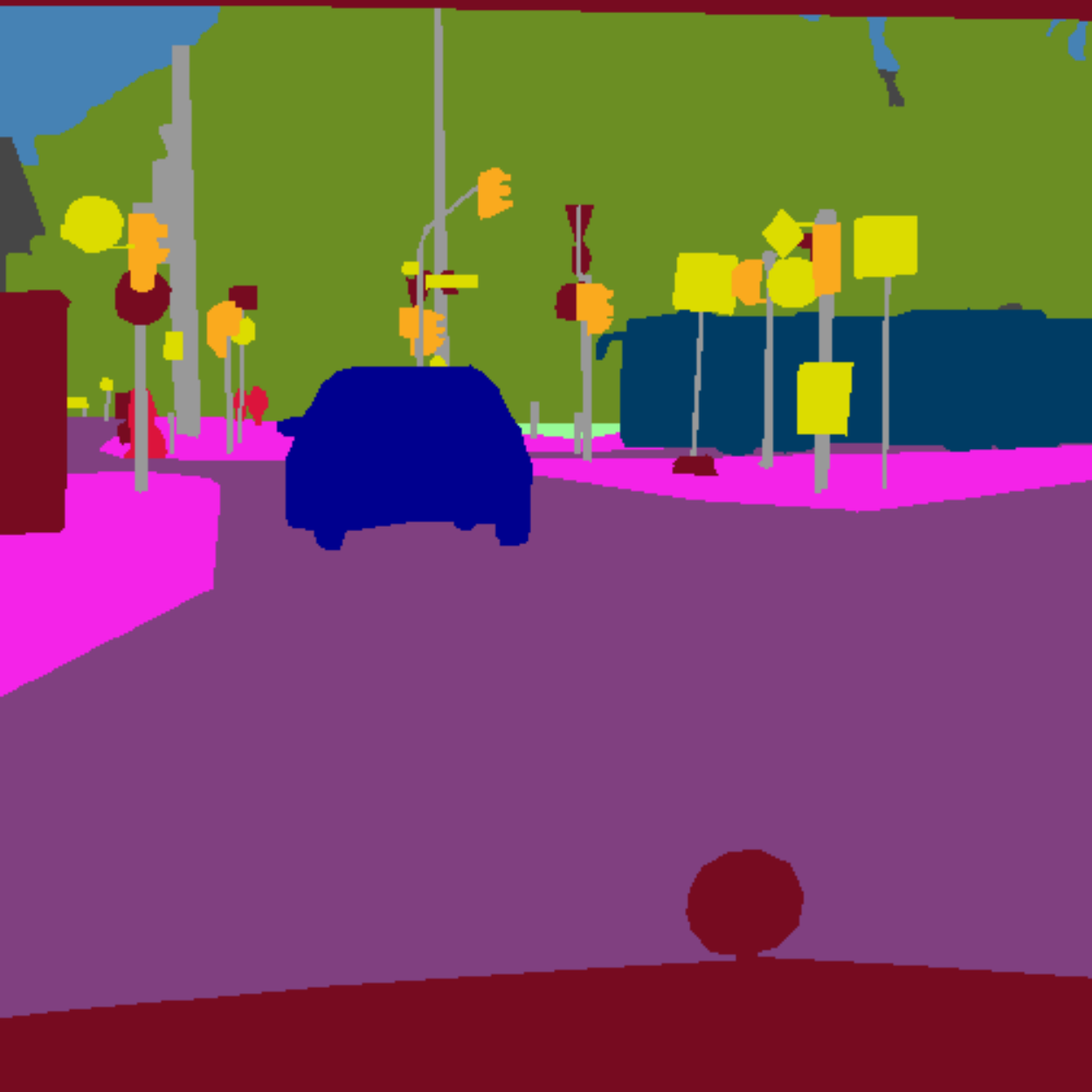} &
        \includegraphics[width=\linewidth]{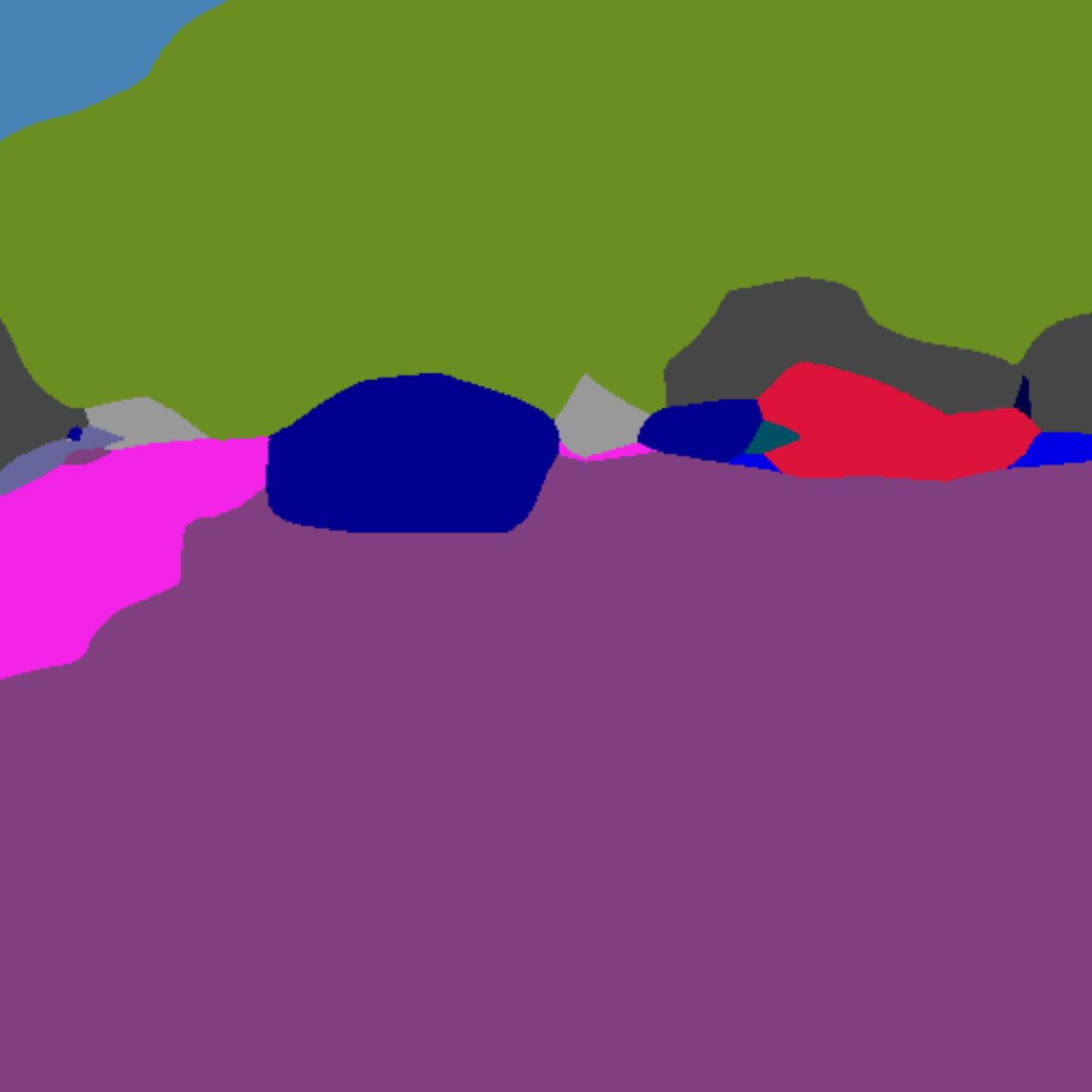} &
        \includegraphics[width=\linewidth]{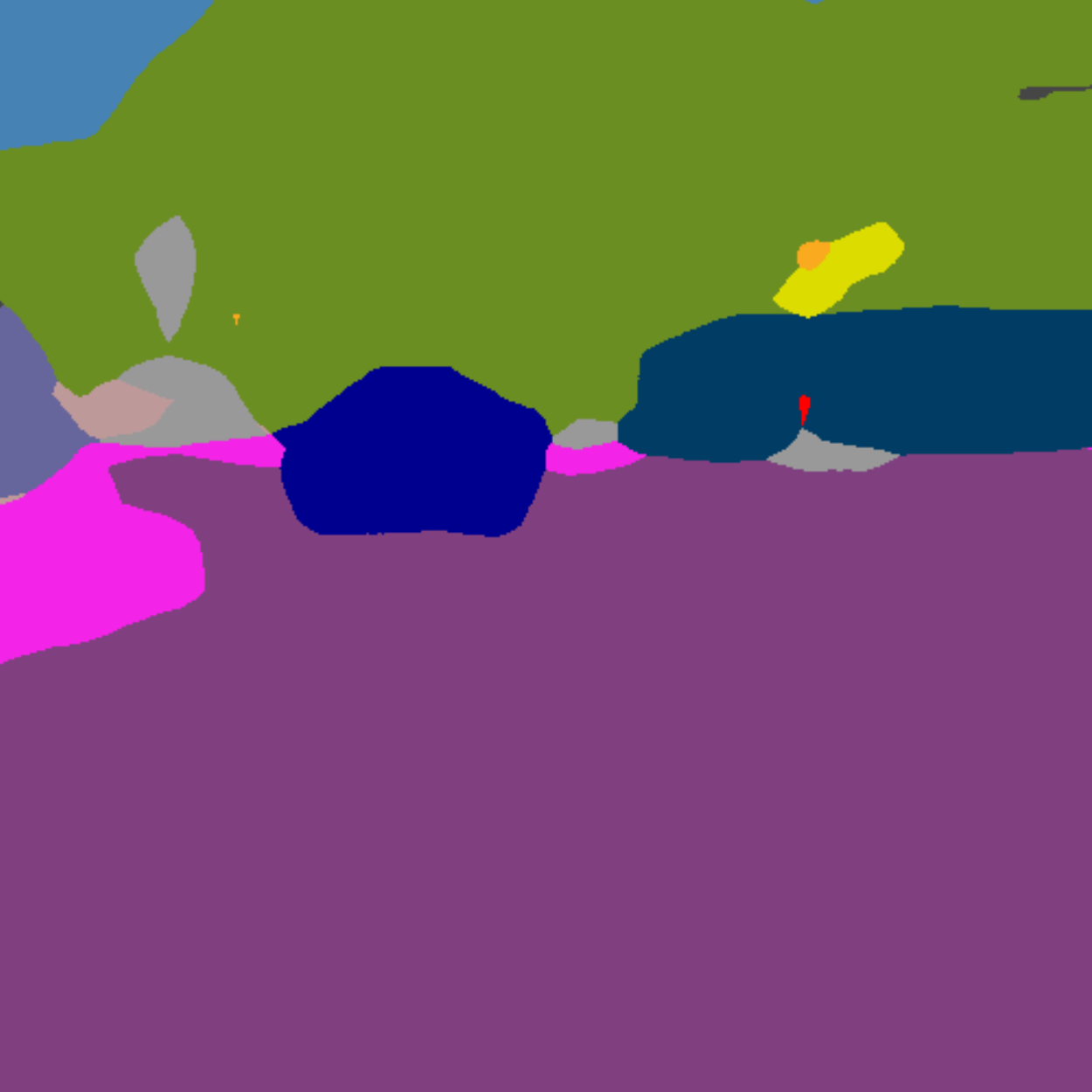} \\
        \includegraphics[width=\linewidth]{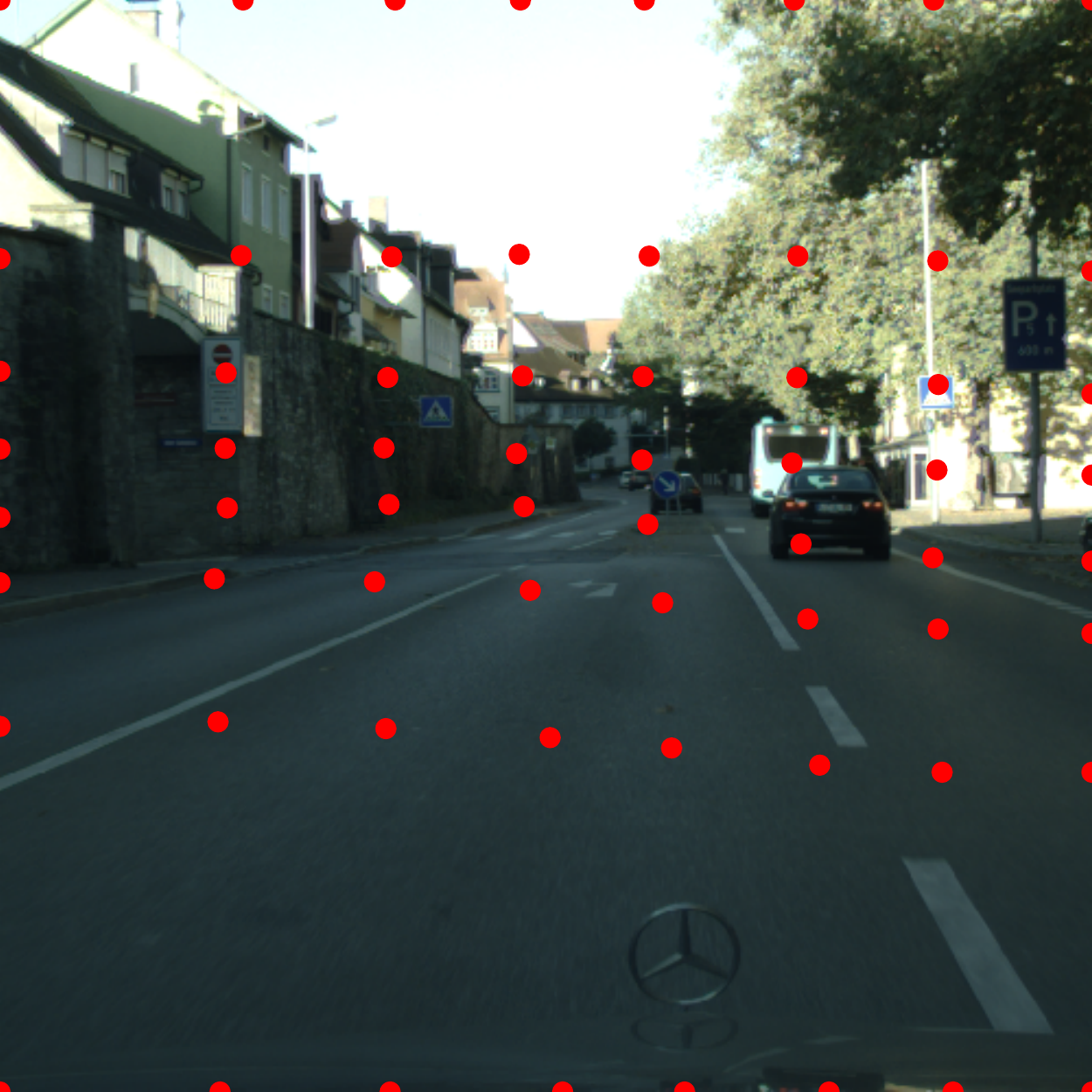} &
        \includegraphics[width=\linewidth]{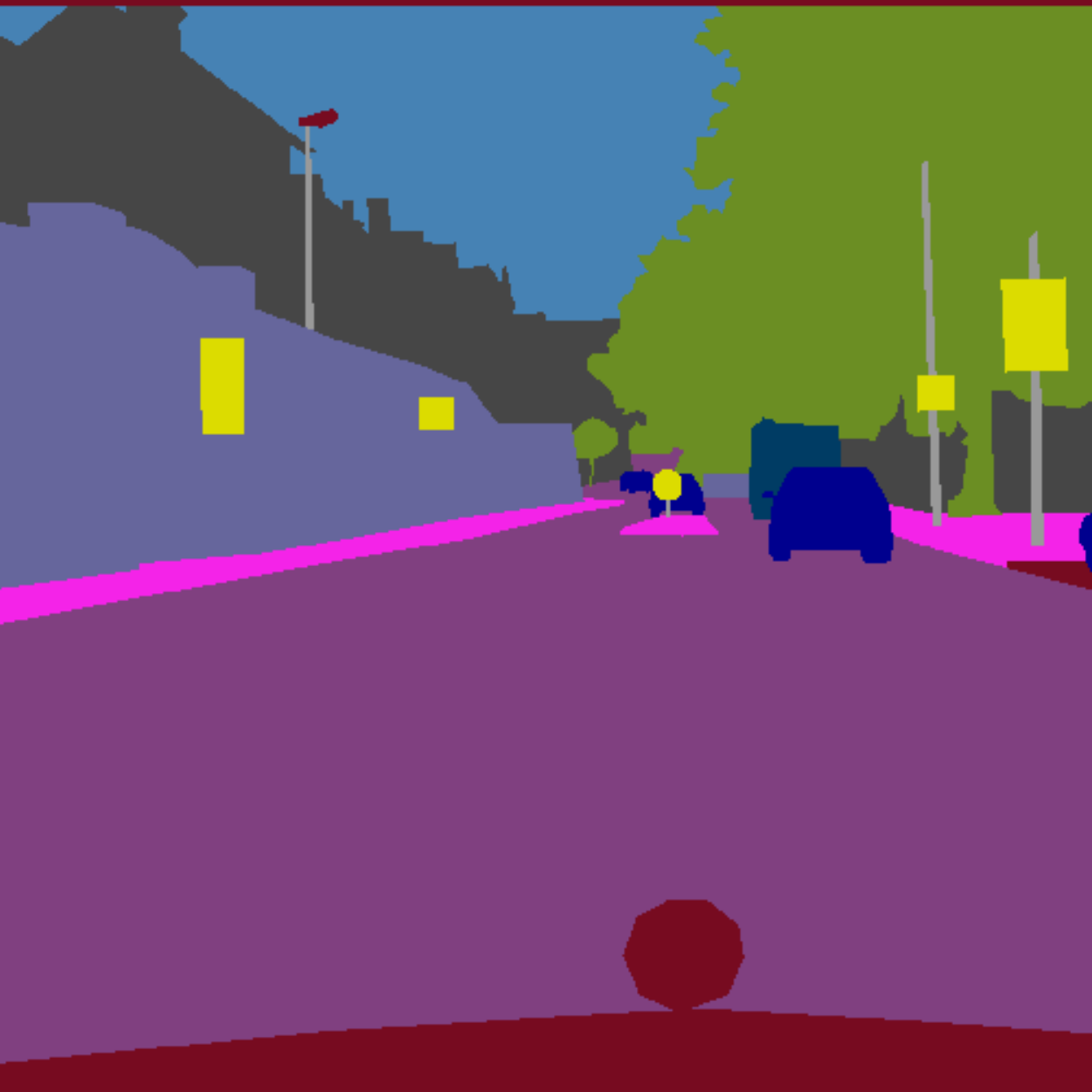} &
        \includegraphics[width=\linewidth]{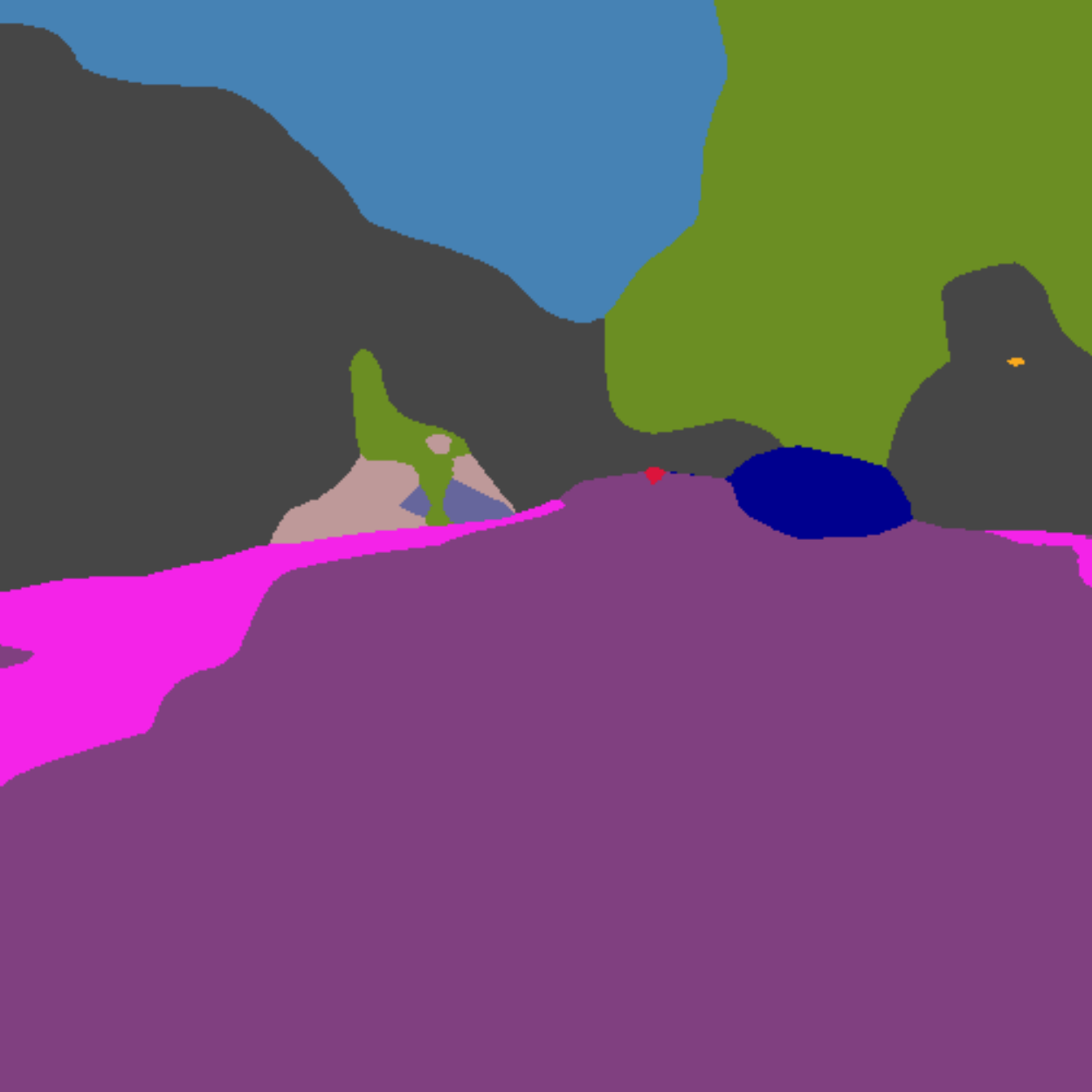} &
        \includegraphics[width=\linewidth]{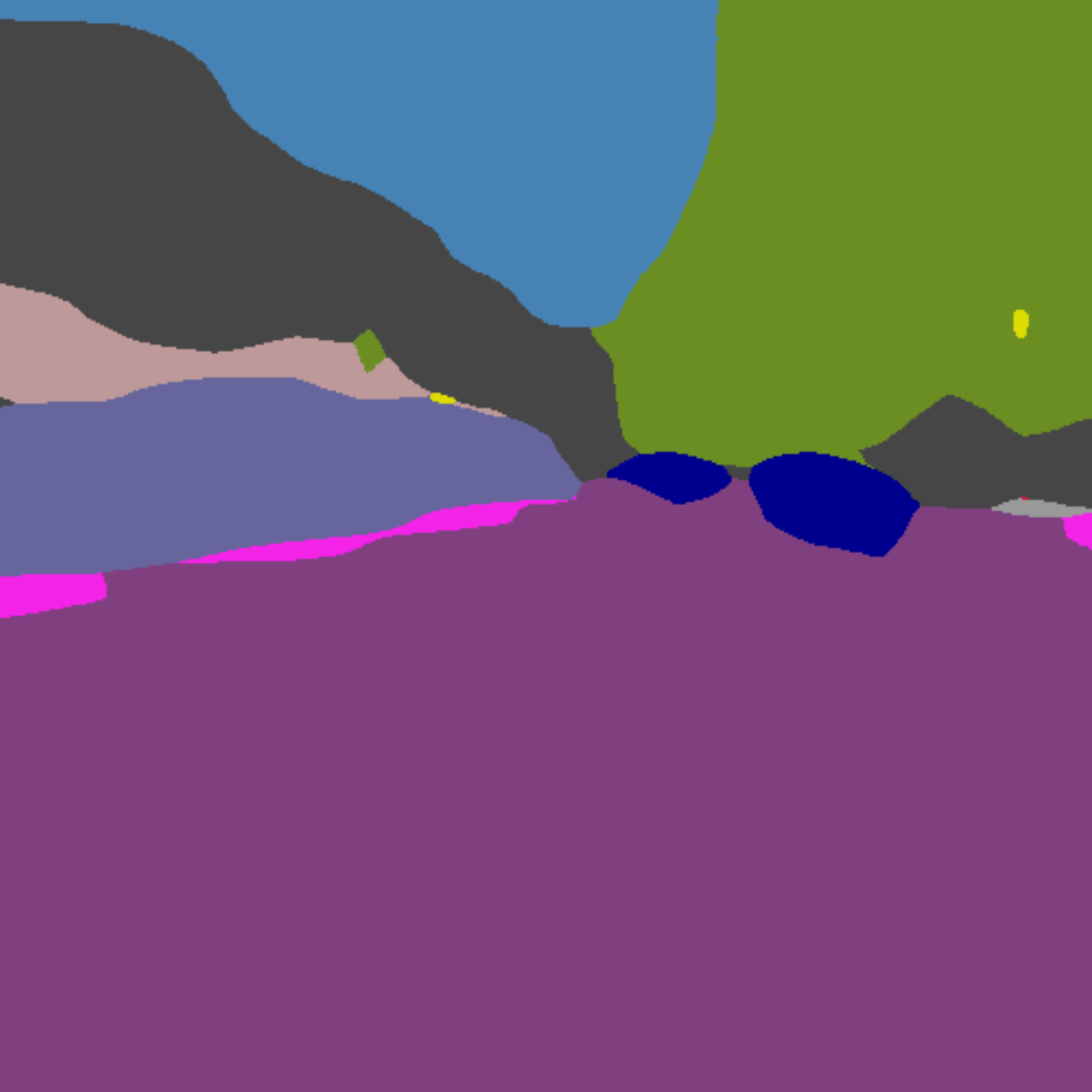} \\
        \includegraphics[width=\linewidth]{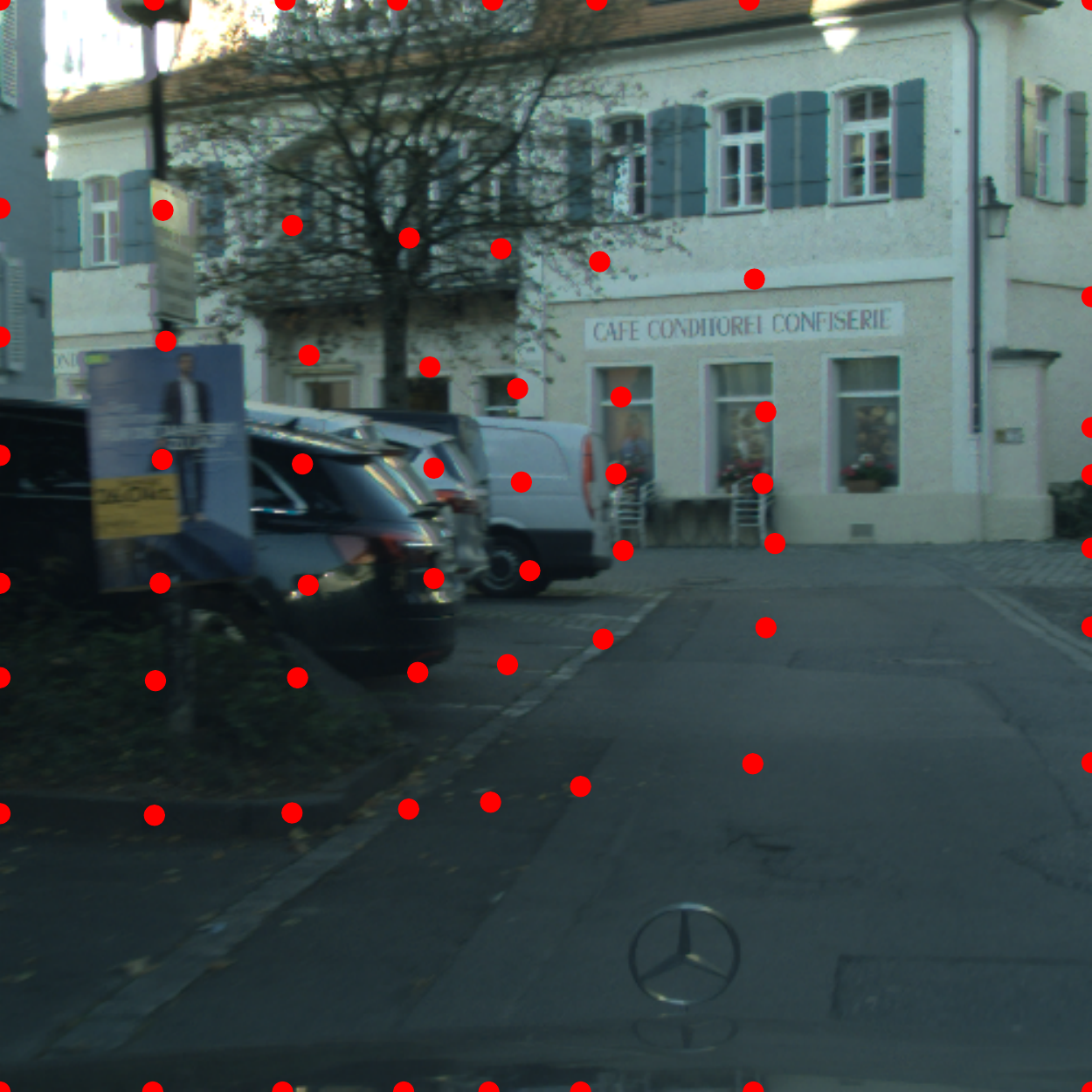} &
        \includegraphics[width=\linewidth]{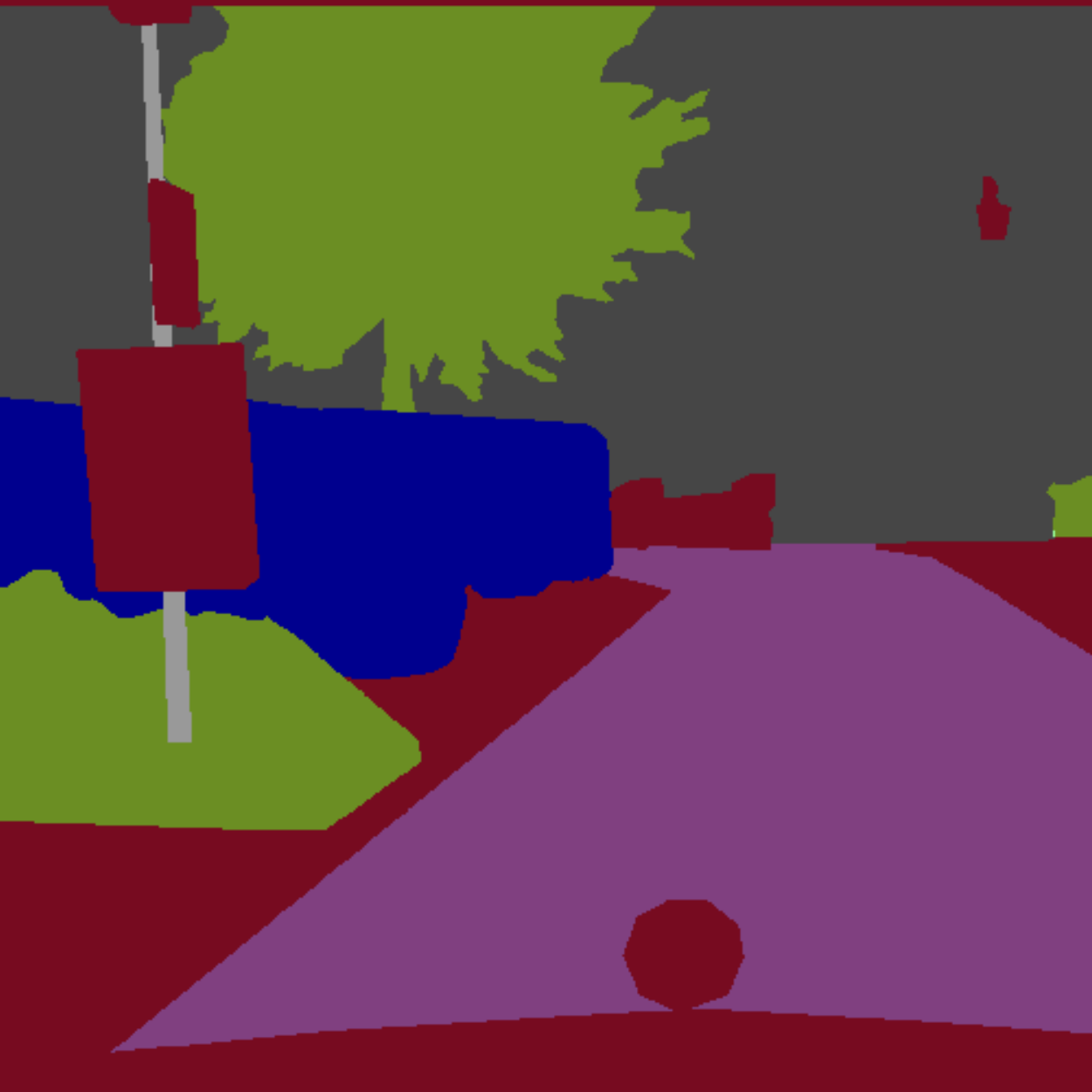} &
        \includegraphics[width=\linewidth]{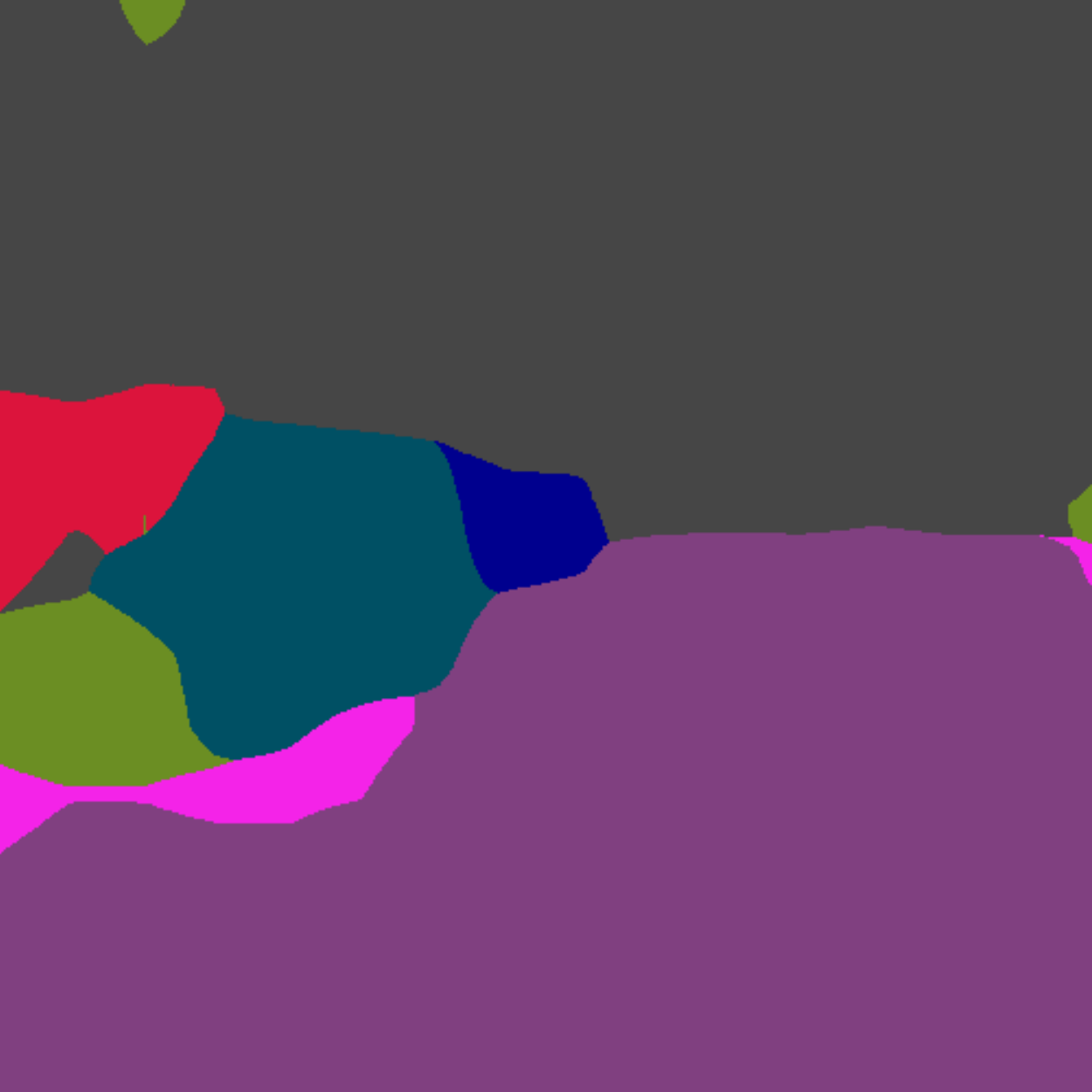} &
        \includegraphics[width=\linewidth]{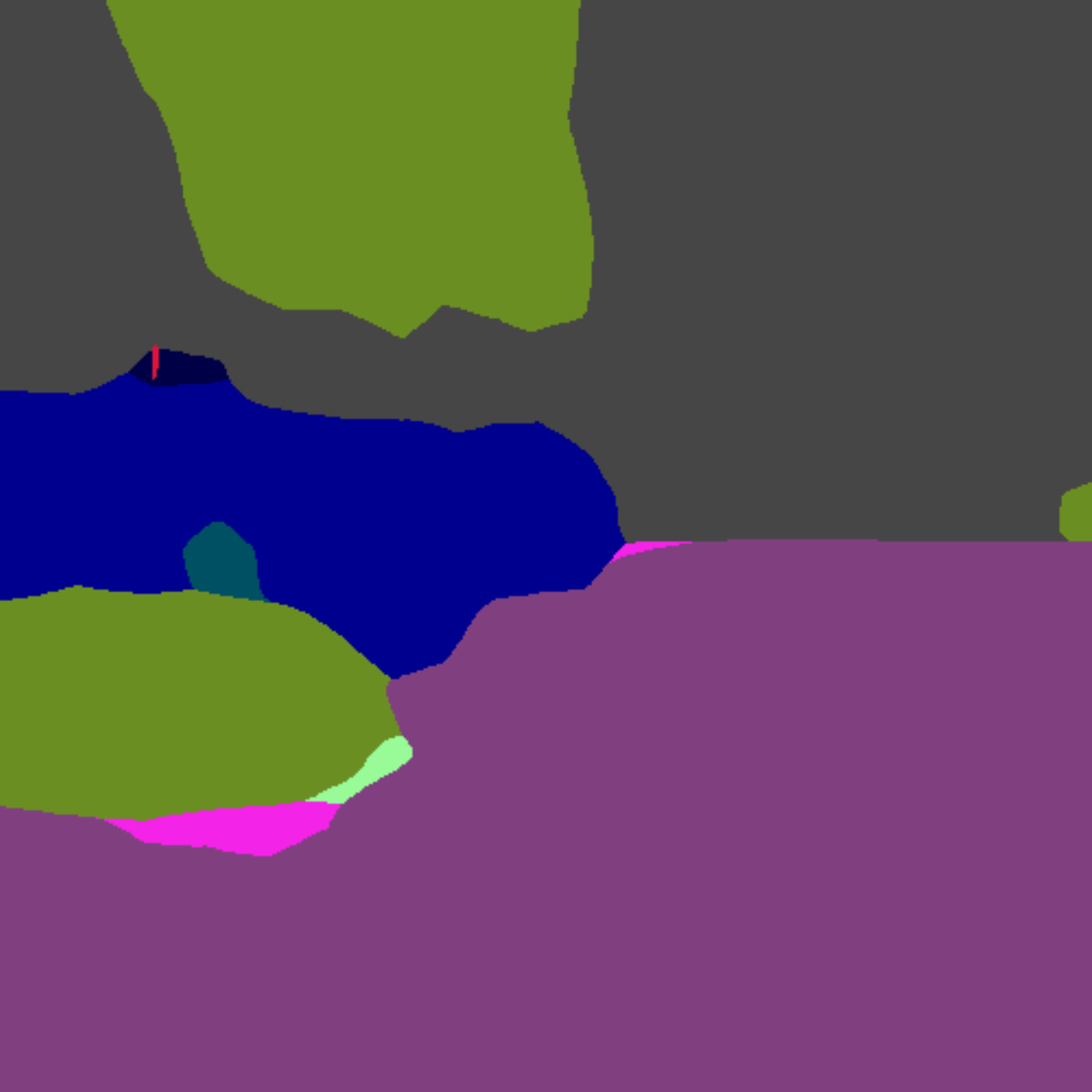}\\
        \includegraphics[width=\linewidth]{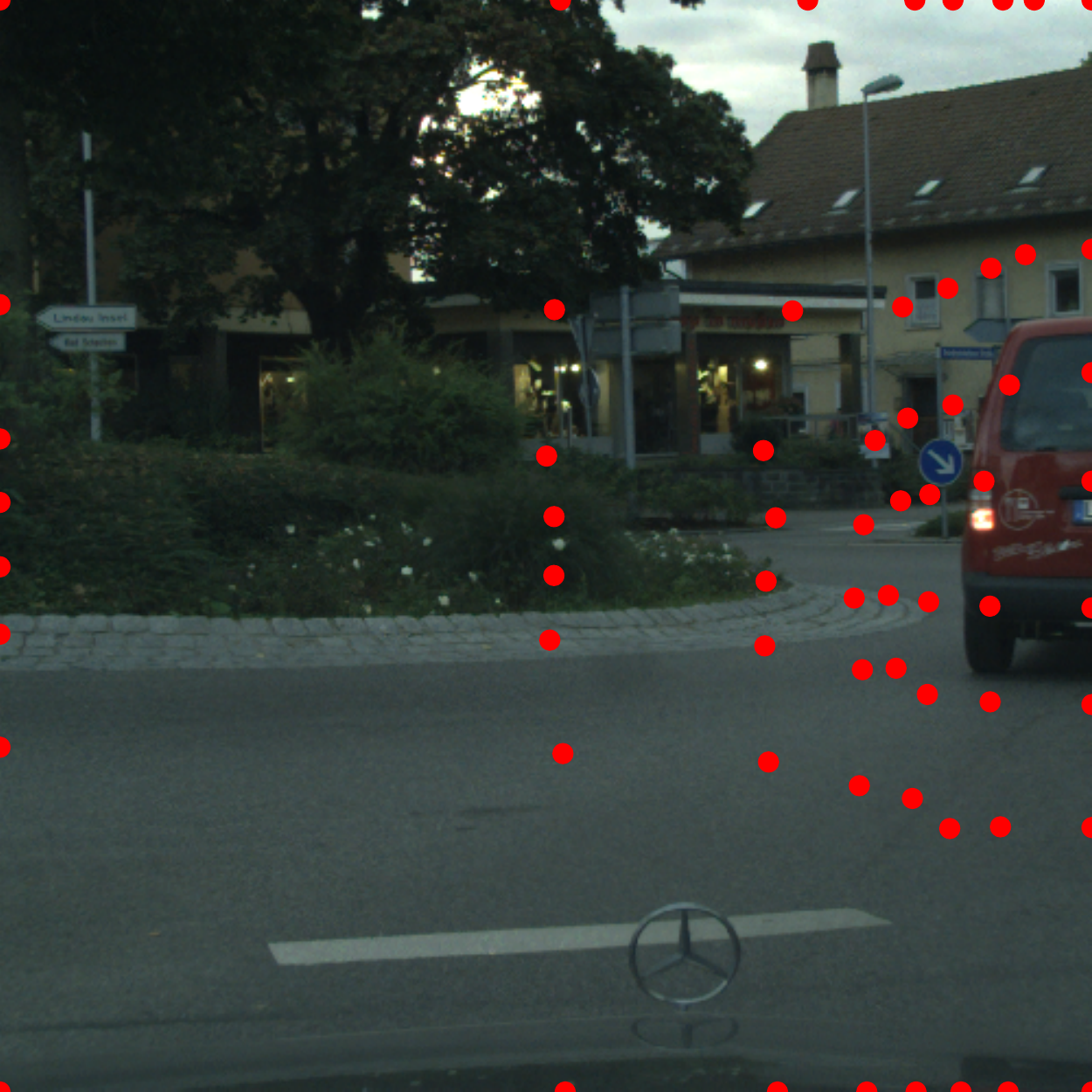} &
        \includegraphics[width=\linewidth]{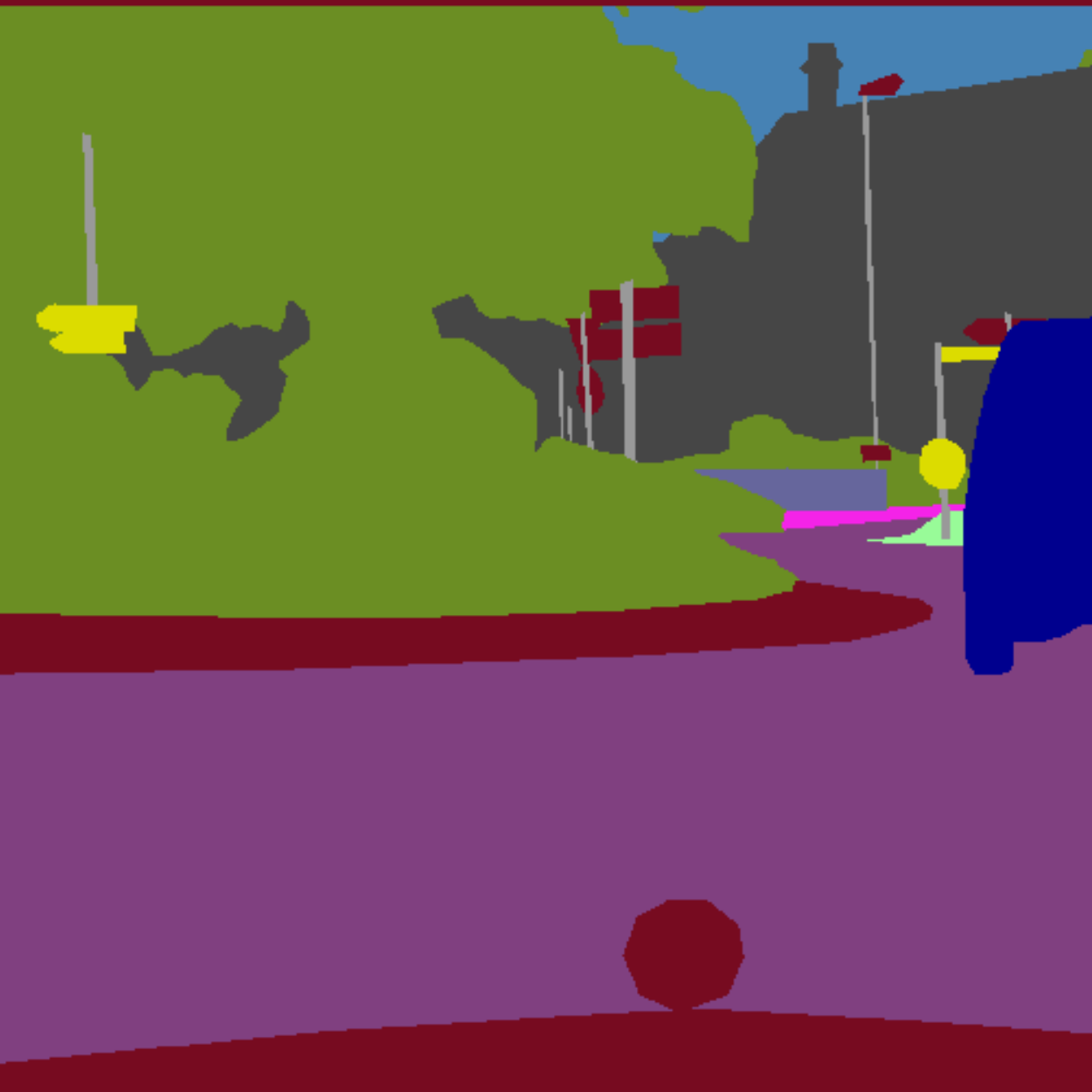} &
        \includegraphics[width=\linewidth]{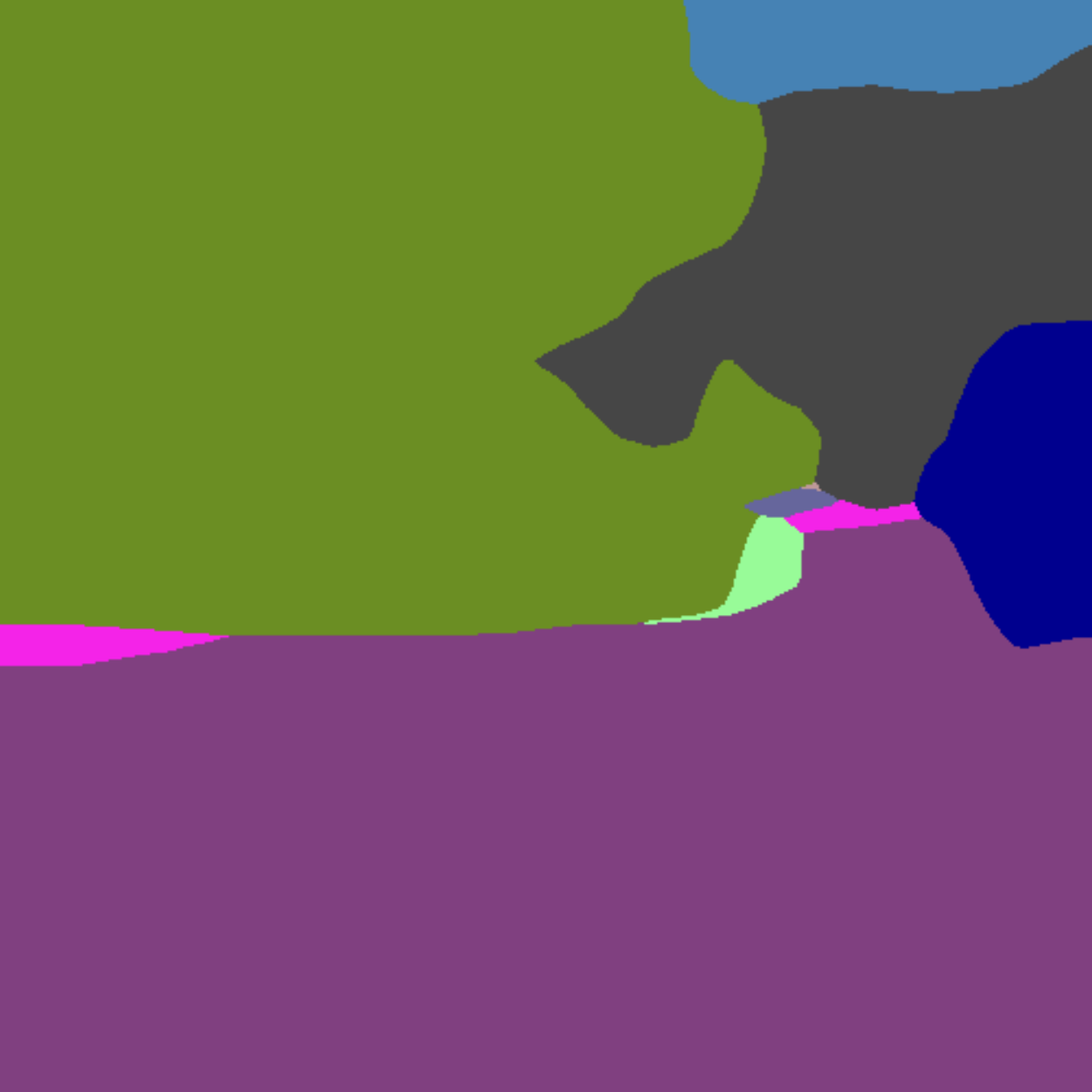} &
        \includegraphics[width=\linewidth]{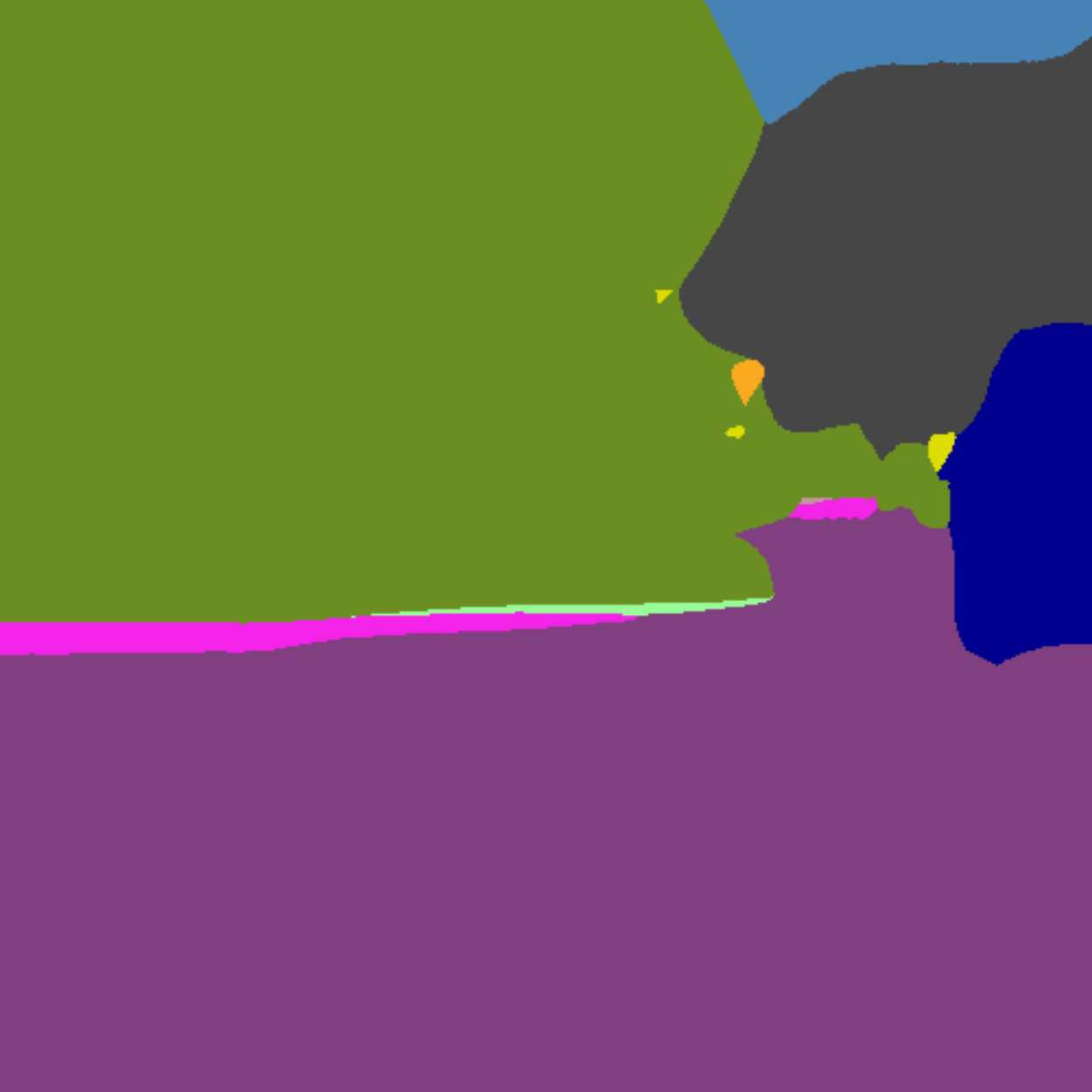}\\
        \includegraphics[width=\linewidth]{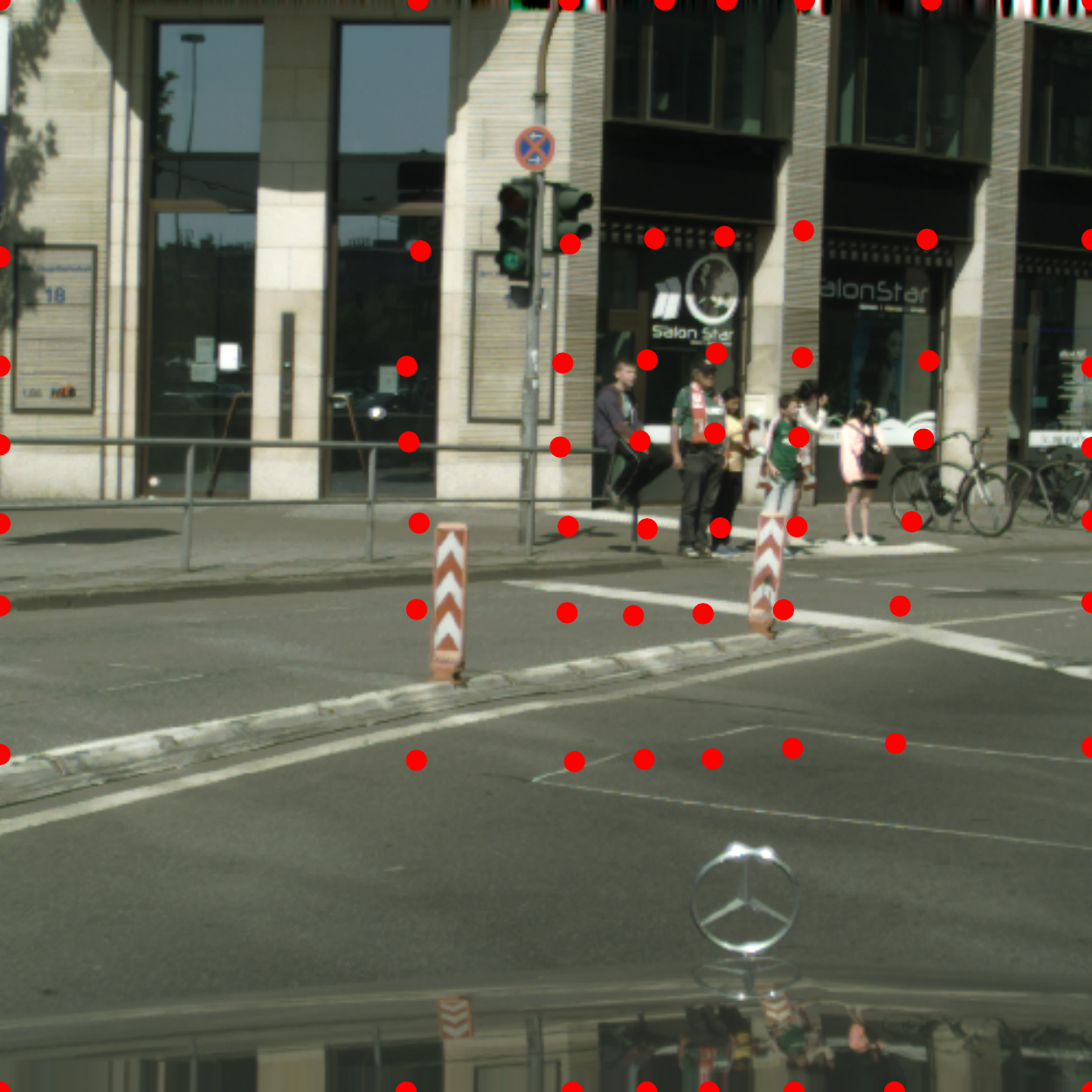} &
        \includegraphics[width=\linewidth]{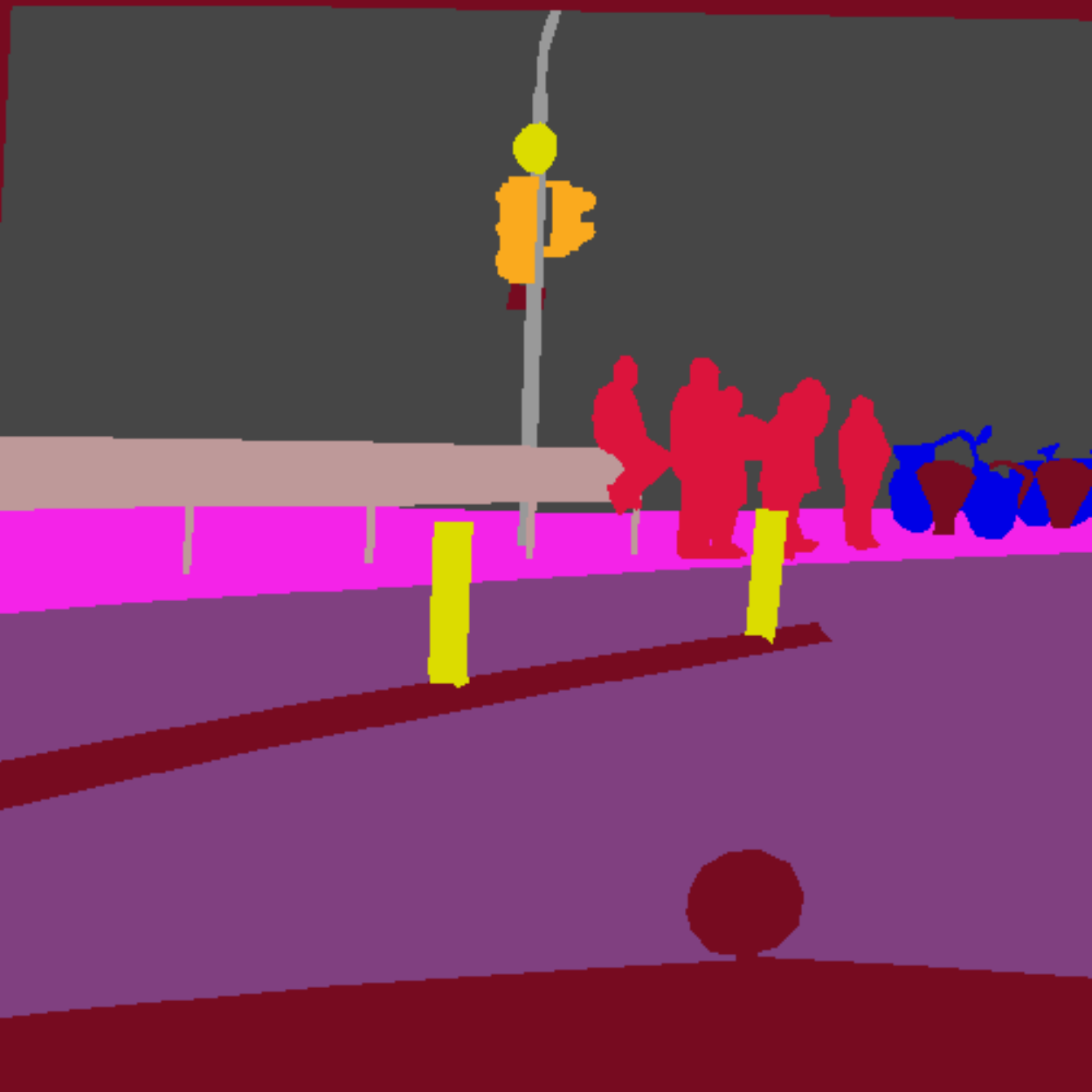} &
        \includegraphics[width=\linewidth]{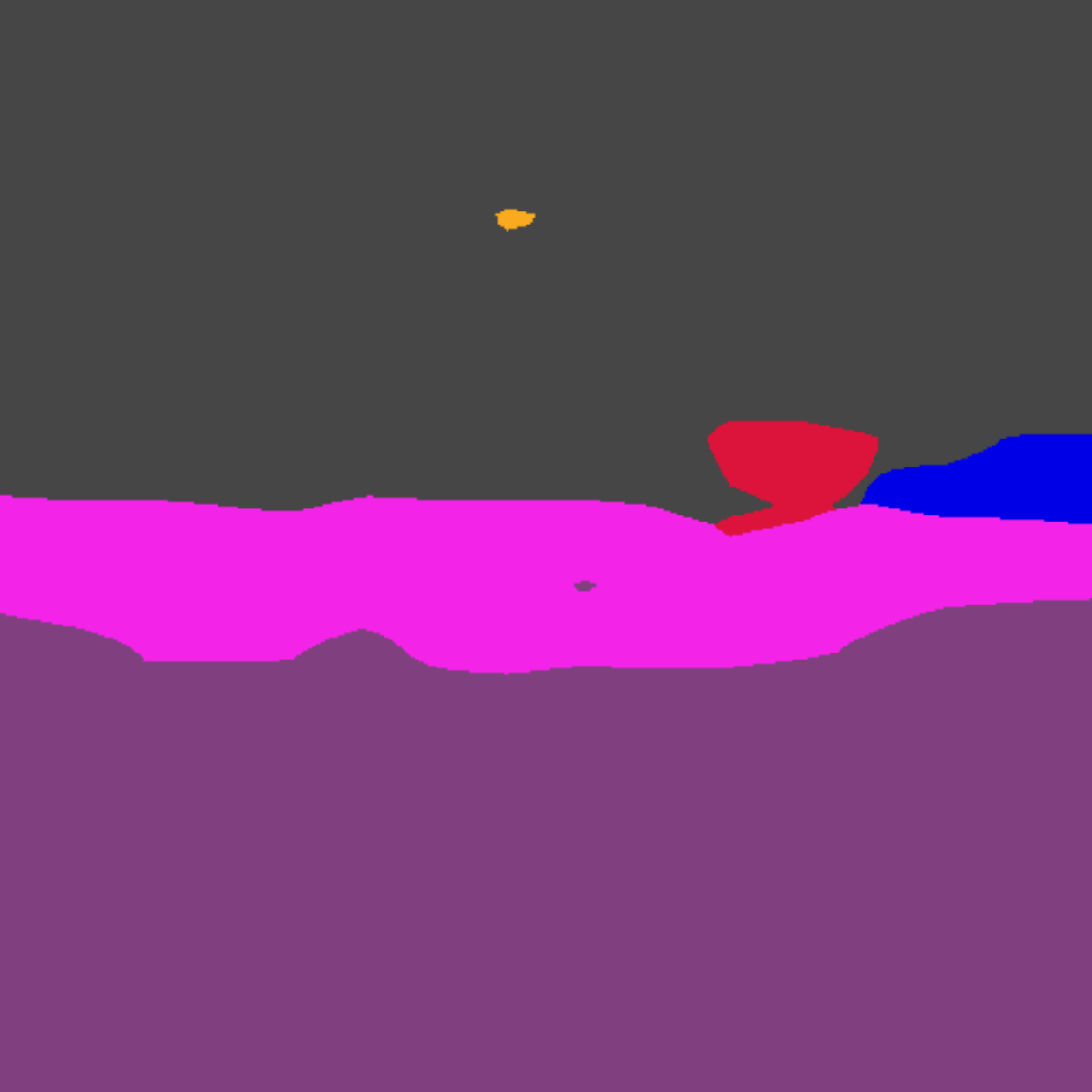} &
        \includegraphics[width=\linewidth]{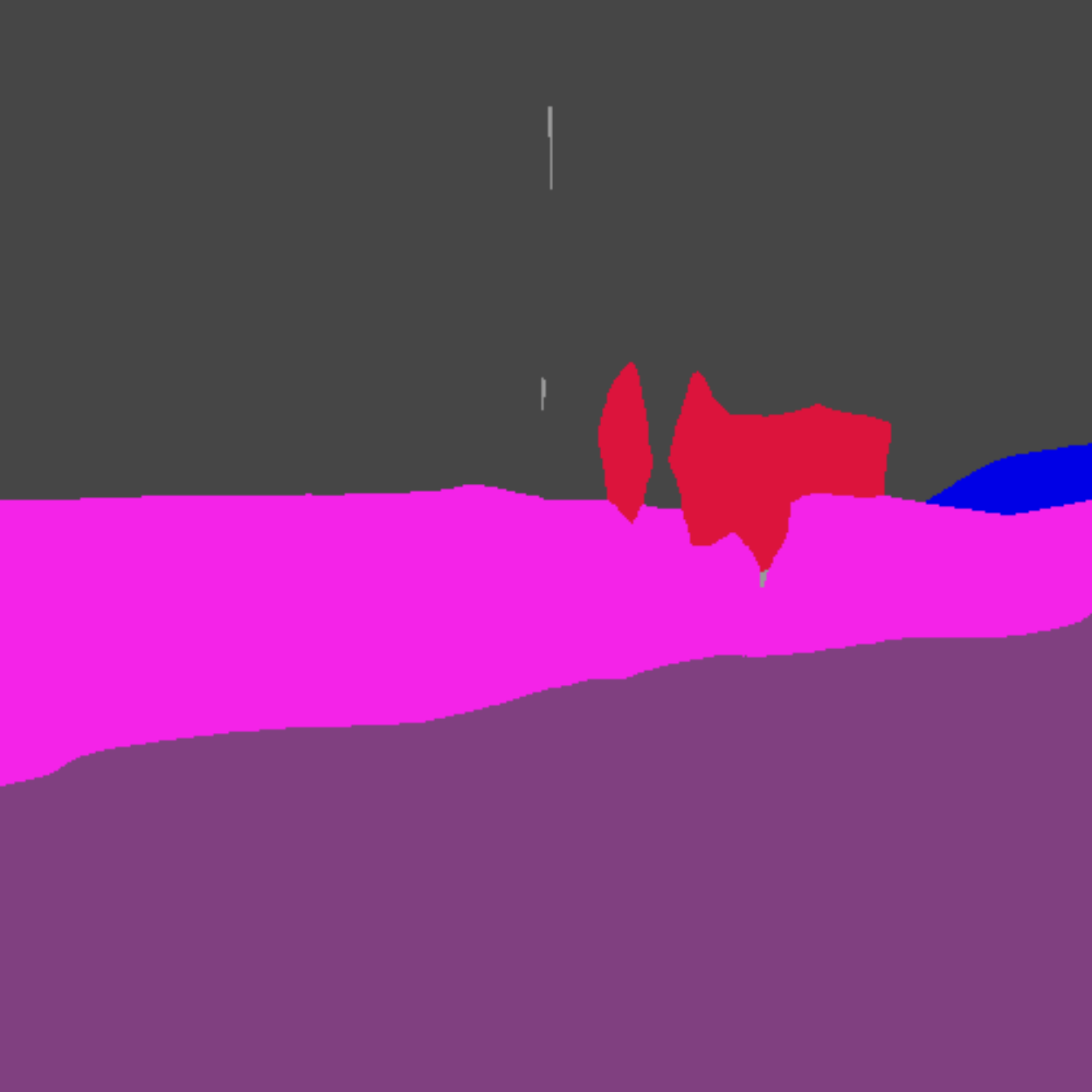} \\
    \end{tabular}
    \vspace{-3mm}
    \caption{Examples from Cityscapes~\cite{Cordts2016Cityscapes} \textit{val} set. \textbf{(a)}: original images and non-uniform $8\times8$ sampling tensor produced by our trained auxiliary net in Fig.~\ref{fig:architecture_sampling} (to avoid clutter, $128\!\times\!128$ tensor interpolation (Fig.~\ref{fig:tensor resizing}) is not shown). \textbf{(c)}: results of the PSP-Net~\cite{zhao2017pyramid} with uniform downsampling to $128 \!\times\! 128$. \textbf{(d)}: results of the same network \textit{with} our adaptive $128 \!\times\! 128$ downsampling based on (a). 
    High-res segmentation results in (c,d) are interpolations (Sec.~\ref{sec:nus:upsample}) of classifications for uniformly and adaptively downsampled pixels.
    }
    \vspace{-1cm}
    \label{fig:qualitative examples}
\end{figure}

In this section we describe several experiments with our adaptive downsampling for semantic segmentation on many high-resolution datasets and state-of-the-art approaches. Figure~\ref{fig:qualitative examples} shows a few qualitative examples.

\subsection{Experimental Setup}
\label{sec:exp:setup}

\paragraph{Dataset and evaluation.} We evaluate and compare the proposed method on several public semantic segmentation datasets. Computational requirements of the contemporaneous approaches and the cost of annotations conditioned the low resolution of images or imprecise (rough) annotations in popular semantic segmentation datasets, such as Caltech~\cite{caltech}, \cite{agarwal2004learning}, Pascal VOC~\cite{pascal-voc-2005,pascal-voc-2012,pascal-voc-2012-aug}, COCO~\cite{coco}. With rapid development of autonomous driving, a number of new semantic segmentation datasets focusing on road scenes~\cite{Cordts2016Cityscapes, apolloscape_2018} or synthetic datasets~\cite{Synthia_2016,Richter_2017} have been made available. These recent datasets provide high-resolution data and high quality annotations. In our experiments, we mainly focus on datasets with high-resolution images, namely ApolloScapes \cite{apolloscape_2018}, CityScapes~\cite{Cordts2016Cityscapes}, Synthia~\cite{Synthia_2016} and Supervisely (person segmentation)~\cite{supervisely} datasets.

The main evaluation metric is mean \emph{Intersection over Union} (mIoU). The metric is always evaluated on segmentation results at the original resolution. We compare performance at various downsampling resolutions to emulate different operating requirements. Occasionally we use other metrics to demonstrate different features of our approach. 

\paragraph{Implementation details:} Our main implementation is in Caffe2~\cite{caffe2}. 
For both the non-uniform sampler network and segmentation network, we use Adam~\cite{kingma2014adam} optimization method with (base learning rate, \#epochs) of ($10^{-5}$, 33), ($10^{-4}$, 1000), ($10^{-4}$, 500) for datasets ApolloScape, Supervisely, and Synthia, respectively. We employ exponential learning rate policy. The batch size is as follows: \\
\begin{tabular}{l|ccccccc}
    input resolution & 16  & 32  & 64  & 128 & 256 & 512  \\ \hline
    batch size       & 128 & 128 & 128 & 32 & 24 & 12
\end{tabular}.

Experiments with PSP-Net~\cite{zhao2017pyramid} and Deeplabv3+~\cite{chen2018encoder} use public implementations with the default parameters.

\begin{figure}[t]
    \centering 
    \begin{tabular}{m{1ex}m{0.9\linewidth}}
    \begin{sideways}\centering U-Net backbone\end{sideways} &
    \includegraphics[width=\linewidth,trim=3mm 12mm 3mm 12mm]{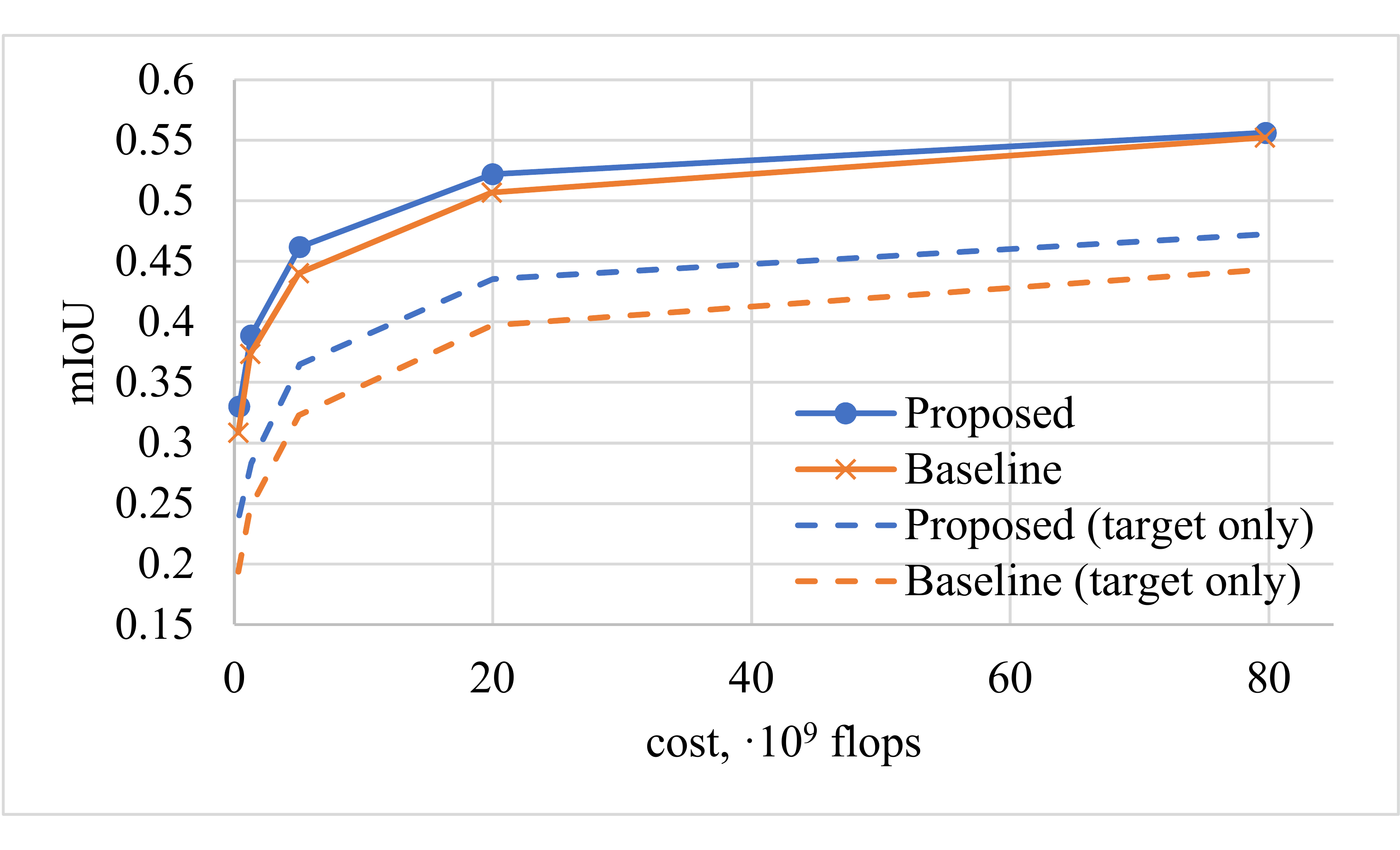}
    \end{tabular}
    \caption{Cost-performance analysis on \textbf{ApolloScape} dataset. Proposed method performs better than the baseline method. With the same cost we can achieve higher quality. 
    }
    \label{fig:apollo cost}
    \vspace{3mm}

    \centering
    \begin{tabular}{m{1ex}m{0.9\linewidth}}
    \begin{sideways}\centering PSP-Net backbone\end{sideways} &
    \includegraphics[width=\linewidth,trim=3mm 12mm 3mm 12mm]{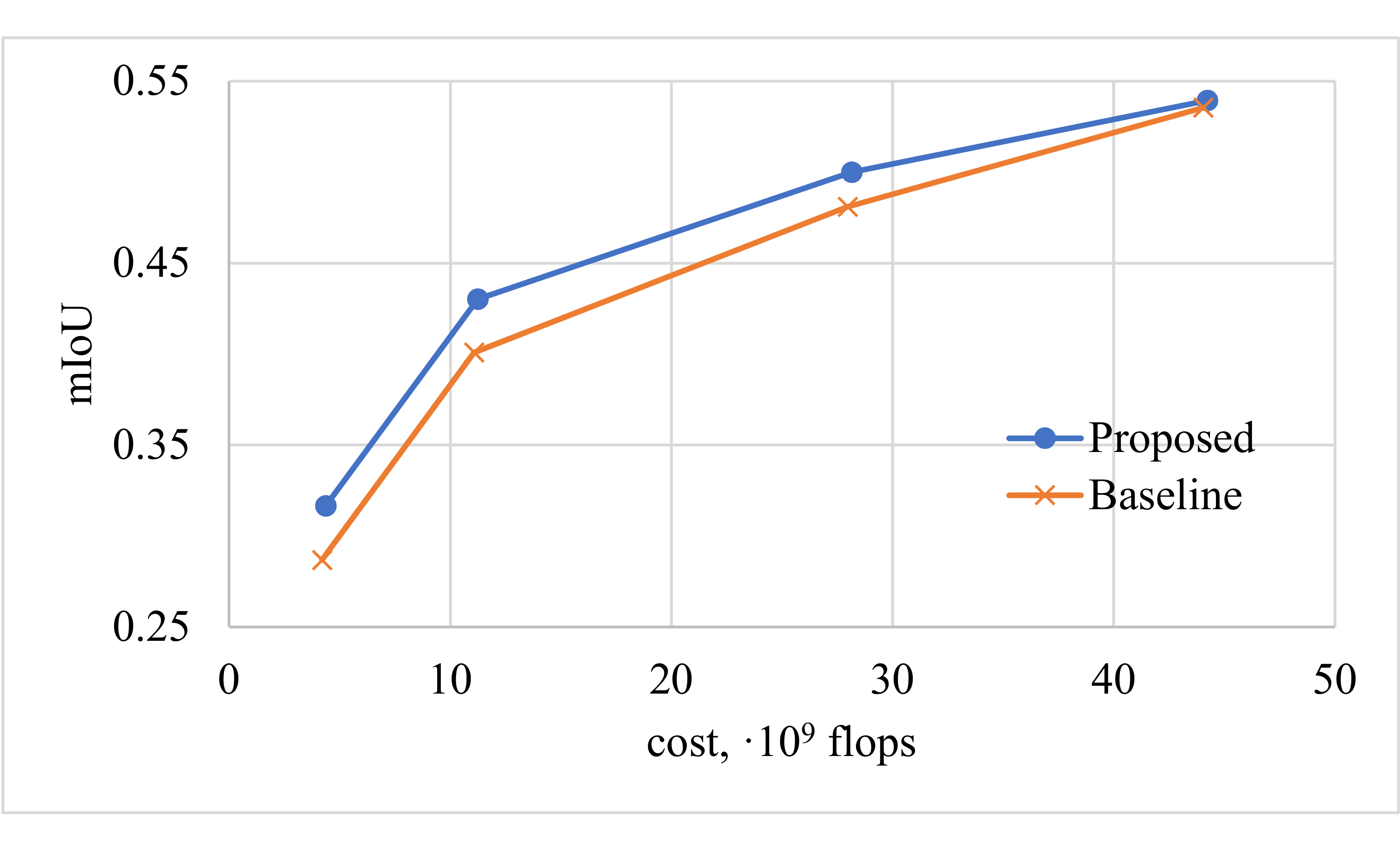}\\[2mm]
    \begin{sideways}\centering Deeplabv3+ backbone\end{sideways} &
    \includegraphics[width=\linewidth,trim=3mm 12mm 3mm 12mm]{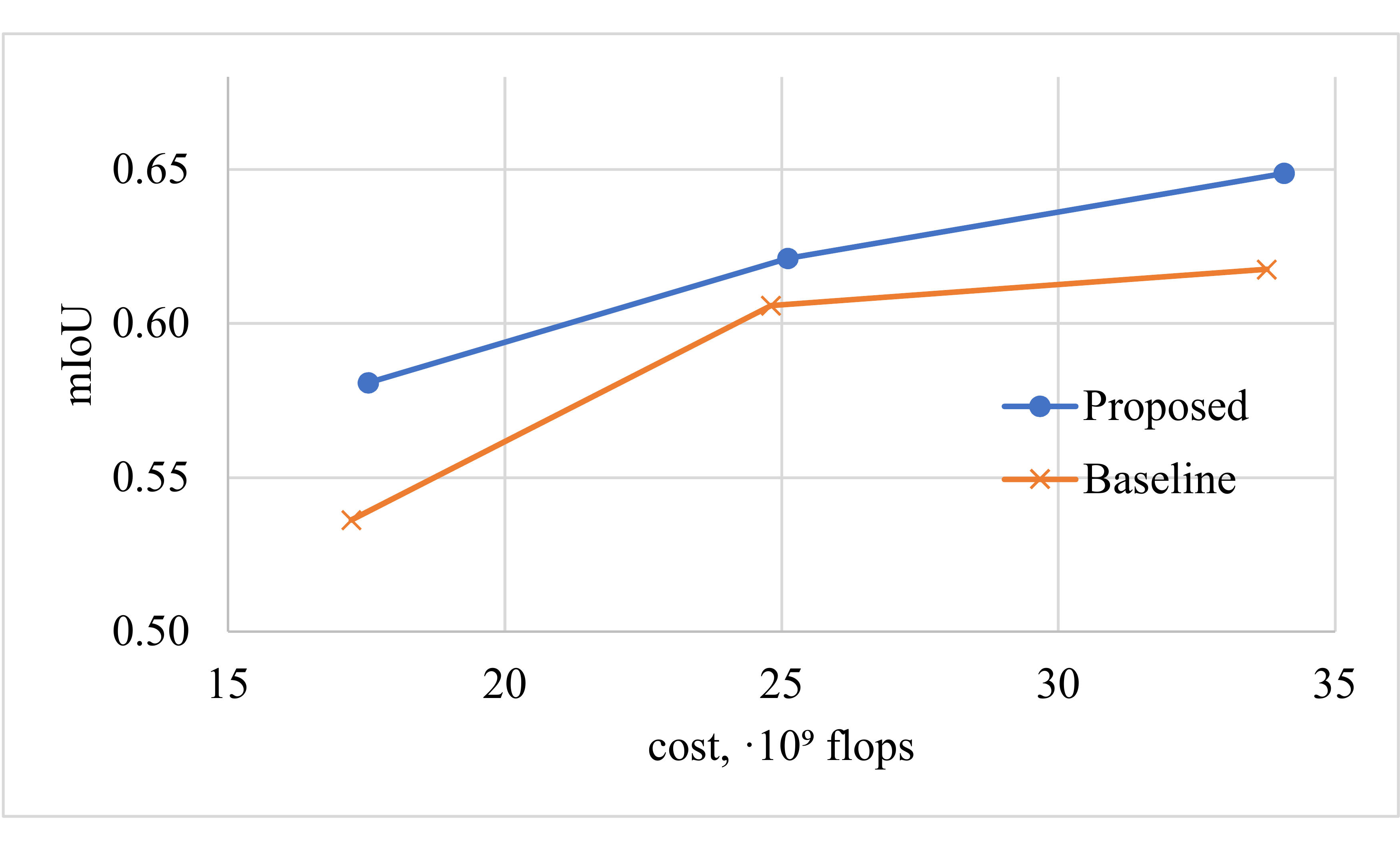}
    \end{tabular}
    \caption{Cost-performance analysis on \textbf{CityScapes} with PSP-Net and Deeplabv3+ baselines for varying downsampling size, see Tab.~\ref{tab:CityScapes results}. Our content-adaptive downsampling gives better results with the same computational cost. }
    \label{fig:cityscapes cost}
\end{figure}

\begin{figure}
    \centering
    \begin{tabular}{m{1ex}m{0.9\linewidth}}
    \begin{sideways}\centering U-Net backbone\end{sideways} &
    \includegraphics[width=\linewidth,trim=3mm 12mm 3mm 12mm]{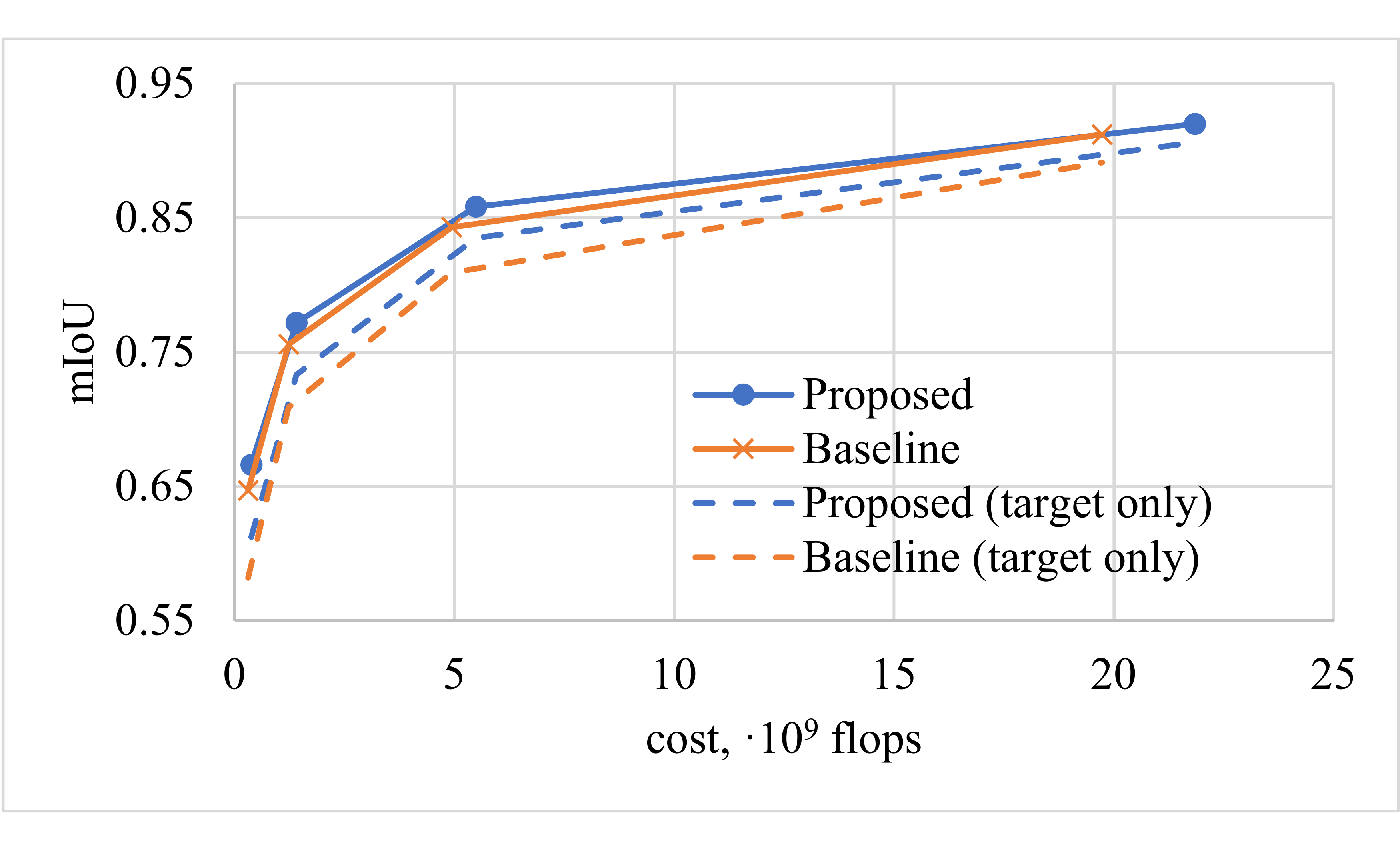}
    \end{tabular}
    \caption{Cost-performance analysis on \textbf{Synthia} dataset. Our approach performs better for target classes (with a tie on all classes).}
    \label{fig:synthia cost}
    \vspace{3mm}

    \centering
    \begin{tabular}{m{1ex}m{0.9\linewidth}}
    \begin{sideways}\centering U-Net backbone\end{sideways} &
    \includegraphics[width=\linewidth,trim=3mm 12mm 3mm 12mm]{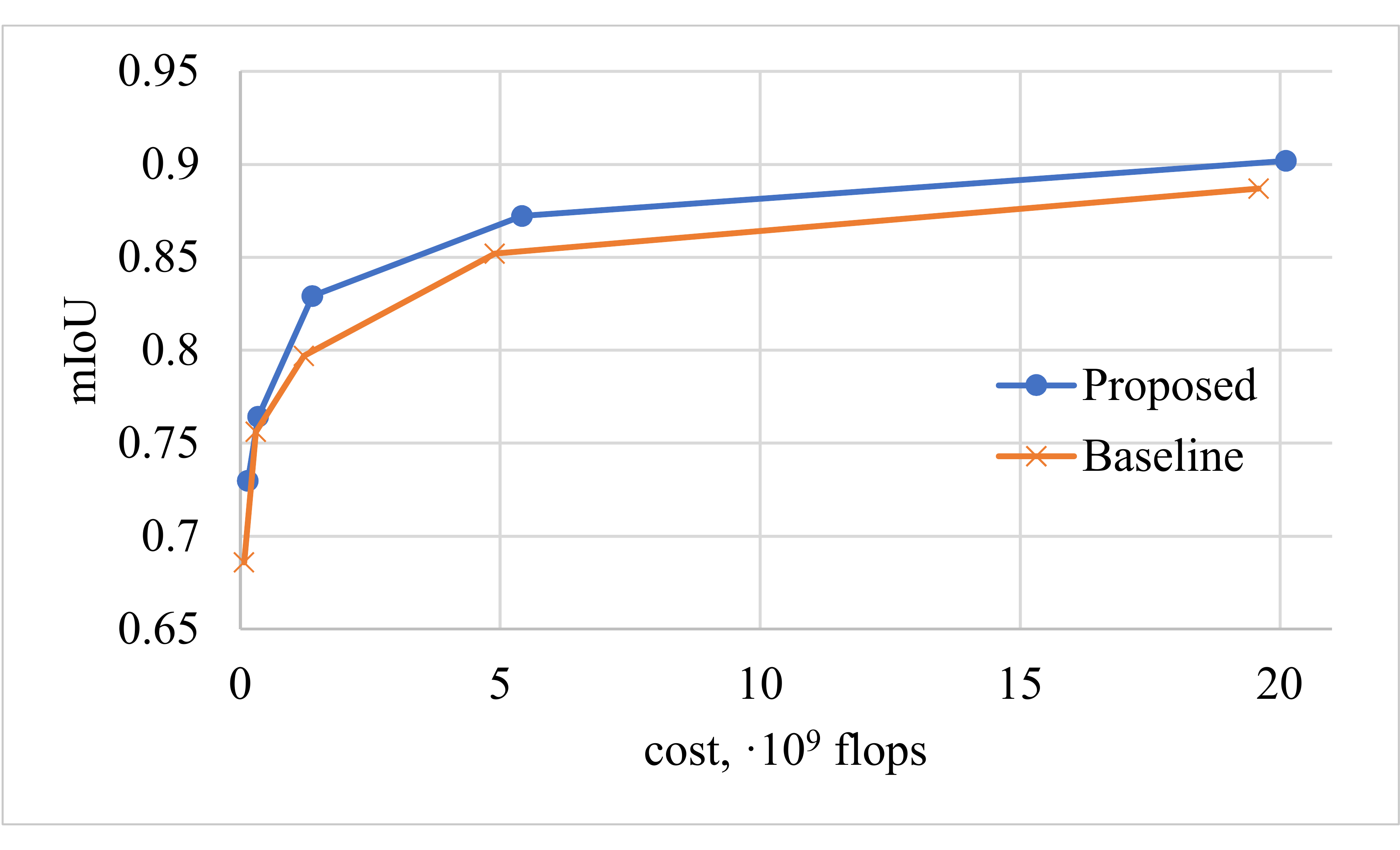}
    \end{tabular}
    \caption{Cost-performance analysis on \textbf{Supervisely} dataset. Our approach improves quality of segmentation.}
    \label{fig:supervisely cost}
\end{figure}

In all experiments, we consider segmentation networks fed with uniformly downsampled images as our baseline. We replace the uniform downsampling with adaptive one as described in Sec.~\ref{sec:nus:sampling_model}. The interpolation of the predictions follows Sec.~\ref{sec:nus:upsample} in both cases. The auxiliary network is separately trained with ground truth produced by~\eqref{eq:energy} where we set $\lambda=1$. The auxiliary network predicts a sampling tensor of size $(2,8,8)$, which is then resized to a required downsampling resolution. During training of the segmentation network we do not include upsampling stage (for both baseline and proposed models) but instead downsample the label map. We use the softmax-entropy loss.

During training we randomly crop largest square from an image. For example, if the original image is $3384\!\times\!2710$ we select a patch of size $2710\!\times\!2710$. During testing we crop the central largest square. Additionally, during training we augment data by random left-right flipping, adjusting the contrast, brightness and adding salt-and-pepper noise.

\begin{table*}[htbp]
\setlength\tabcolsep{2pt}\footnotesize
  \centering
    \makebox[\textwidth][c]{
    \begin{tabular}{l|cc|cccccccccccccc|cccccccc|cc}
      & \multicolumn{1}{c}{\multirow{2}[1]{*}{\begin{sideways}\parbox{15mm}{downsample\\[-0.8ex] resolution}\end{sideways}}} & \multirow{2}[1]{*}{\begin{sideways}\parbox{15mm}{flops, $\cdot 10^9$}\end{sideways}} & \multicolumn{14}{c|}{non-target classes, IoU}              & \multicolumn{8}{c|}{\textbf{target classes}, IoU} & \multicolumn{2}{c}{mIoU} \\
      &   &   & {\begin{sideways}road\end{sideways}} & {\begin{sideways}sidewalk\end{sideways}} & {\begin{sideways}\parbox{10mm}{traffic\\[-0.8ex]cone}\end{sideways}} & {\begin{sideways}\parbox{10mm}{road pile}\end{sideways}} & {\begin{sideways}fence\end{sideways}} & {\begin{sideways}\parbox{10mm}{traffic\\[-0.8ex] light}\end{sideways}} & {\begin{sideways}pole\end{sideways}} & {\begin{sideways}\parbox{10mm}{traffic\\[-0.8ex]sign}\end{sideways}} & {\begin{sideways}wall\end{sideways}} & {\begin{sideways}dustbin\end{sideways}} & {\begin{sideways}billboard\end{sideways}} & {\begin{sideways}building\end{sideways}} & {\begin{sideways}vegatation\end{sideways}} & {\begin{sideways}sky\end{sideways}} & {\begin{sideways}\textbf{car}\end{sideways}} & {\begin{sideways}\parbox{10mm}{\textbf{motor-\\[-0.8ex]bicycle}}\end{sideways}} & {\begin{sideways}\textbf{bicycle}\end{sideways}} & {\begin{sideways}\textbf{person}\end{sideways}} & {\begin{sideways}\textbf{rider}\end{sideways}} & {\begin{sideways}\textbf{truck}\end{sideways}} & {\begin{sideways}\textbf{bus}\end{sideways}} & {\begin{sideways}\textbf{tricycle}\end{sideways}} & {\begin{sideways}\parbox{10mm}{all\\[-0.8ex]classes}\end{sideways}} & {\begin{sideways}\parbox{10mm}{\textbf{target\\[-0.8ex] classes}}\end{sideways}} \\
    \hline
    \textbf{Ours} & 32 & 0.38 & \textbf{0.92} & \textbf{0.38} & \textbf{0.17} & \textbf{0.00} & \textbf{0.49} & 0.11 & 0.08 & 0.44 & \textbf{0.28} & \textbf{0.03} & \textbf{0.00} & 0.74 & 0.86 & 0.84 & \textbf{0.66} & \textbf{0.07} & \textbf{0.27} & \textbf{0.02} & \textbf{0.03} & \textbf{0.34} & \textbf{0.52} & \textbf{0.01} & \textbf{0.24} & \textbf{0.24} \\
    \textbf{Baseline} & 32 & \textbf{0.31} & 0.92 & 0.29 & 0.13 & 0.00 & 0.43 & \textbf{0.14} & \textbf{0.11} & \textbf{0.53} & 0.18 & 0.00 & 0.00 & \textbf{0.74} & \textbf{0.87} & \textbf{0.89} & 0.59 & 0.04 & 0.26 & 0.01 & 0.02 & 0.20 & 0.44 & 0.00 & 0.19 & 0.19 \\
    \hline
    \textbf{Ours} & 64 & 1.31 & 0.94 & 0.39 & \textbf{0.31} & \textbf{0.02} & \textbf{0.56} & 0.25 & 0.17 & 0.61 & \textbf{0.41} & \textbf{0.08} & \textbf{0.00} & 0.78 & 0.89 & 0.87 & \textbf{0.76} & \textbf{0.10} & \textbf{0.33} & \textbf{0.04} & \textbf{0.03} & \textbf{0.44} & \textbf{0.53} & \textbf{0.04} & \textbf{0.28} & \textbf{0.28} \\
    \textbf{Baseline} & 64 & \textbf{1.24} & \textbf{0.94} & \textbf{0.40} & 0.30 & 0.01 & 0.52 & \textbf{0.30} & \textbf{0.22} & \textbf{0.64} & 0.29 & 0.04 & 0.00 & \textbf{0.79} & \textbf{0.90} & \textbf{0.91} & 0.70 & 0.06 & 0.31 & 0.02 & 0.03 & 0.32 & 0.52 & 0.03 & 0.25 & 0.25 \\
    \hline
    \textbf{Ours} & 128 & 5.05 & 0.95 & \textbf{0.51} & \textbf{0.43} & \textbf{0.07} & \textbf{0.61} & 0.44 & 0.29 & 0.71 & \textbf{0.47} & \textbf{0.13} & \textbf{0.01} & 0.82 & 0.91 & 0.88 & \textbf{0.83} & \textbf{0.16} & \textbf{0.41} & \textbf{0.08} & \textbf{0.05} & \textbf{0.57} & \textbf{0.76} & 0.06 & \textbf{0.36} & \textbf{0.36} \\
    \textbf{Baseline} & 128 & \textbf{4.98} & \textbf{0.96} & 0.39 & 0.43 & 0.05 & 0.59 & \textbf{0.45} & \textbf{0.36} & \textbf{0.73} & 0.37 & 0.11 & 0.00 & \textbf{0.83} & \textbf{0.92} & \textbf{0.93} & 0.80 & 0.10 & 0.38 & 0.06 & 0.03 & 0.44 & 0.70 & \textbf{0.06} & 0.32 & 0.32 \\
    \hline
    \textbf{Ours} & 256 & 19.99 & 0.96 & 0.44 & \textbf{0.51} & \textbf{0.13} & \textbf{0.66} & \textbf{0.58} & 0.42 & 0.78 & \textbf{0.58} & \textbf{0.27} & \textbf{0.00} & 0.84 & 0.92 & 0.89 & \textbf{0.88} & \textbf{0.21} & \textbf{0.47} & \textbf{0.18} & 0.04 & \textbf{0.65} & 0.80 & \textbf{0.24} & \textbf{0.44} & \textbf{0.44} \\
    \textbf{Baseline} & 256 & \textbf{19.92} & \textbf{0.97} & \textbf{0.48} & 0.49 & 0.13 & 0.64 & 0.58 & \textbf{0.46} & \textbf{0.79} & 0.48 & 0.24 & 0.00 & \textbf{0.85} & \textbf{0.94} & \textbf{0.94} & 0.86 & 0.17 & 0.42 & 0.15 & \textbf{0.04} & 0.60 & \textbf{0.83} & 0.10 & 0.40 & 0.40 \\
    \hline
    \textbf{Ours} & 512 & 79.76 & 0.97 & 0.44 & 0.54 & \textbf{0.21} & 0.68 & 0.63 & 0.49 & 0.80 & \textbf{0.67} & 0.36 & 0.00 & 0.85 & 0.93 & 0.90 & \textbf{0.91} & \textbf{0.24} & \textbf{0.52} & \textbf{0.30} & \textbf{0.06} & \textbf{0.75} & 0.81 & \textbf{0.19} & \textbf{0.47} & \textbf{0.47} \\
    \textbf{Baseline} & 512 & \textbf{79.68} & \textbf{0.97} & \textbf{0.47} & \textbf{0.55} & 0.20 & \textbf{0.68} & \textbf{0.67} & \textbf{0.54} & \textbf{0.83} & 0.59 & \textbf{0.36} & \textbf{0.00} & \textbf{0.87} & \textbf{0.94} & \textbf{0.94} & 0.90 & 0.21 & 0.49 & 0.26 & 0.03 & 0.68 & \textbf{0.84} & 0.13 & 0.44 & 0.44 \\[-1ex]
    \bottomrule
    \end{tabular}}%
    \vspace{-2ex}
  \caption{Per class results on the validation set of ApolloScape. Our adaptive sampling improves overall quality of segmentation. Target classes (bold font on the top row) consistently benefit for all resolutions. }
  \label{tab:apollo_per_class}%
\end{table*}


\subsection{Cost-performance Analysis}\label{sec:exp:apollo}


\paragraph{ApolloScape~\cite{apolloscape_2018}} is an open dataset for autonomous driving. The dataset consists of approximately 105K training and 8K validation images of size $3384\!\times\!2710$. The annotations contain 22 classes for evaluation. The annotations of some classes (cars, motorbikes, bicycles, persons, riders, trucks, buses and tricycles) are of high quality. These occupy $26\%$ of pixels in evaluation set. We refer to these as \emph{target classes}. Other classes annotations are noisy. Since the noise in pixel labels greatly magnifies the noise of segments boundaries, we chose to define our sampling model based on the \emph{target} classes boundaries. This exploits an important aspect of our method, \ie an ability to focus on boundaries of specific semantic classes of interest. Following~\cite{apolloscape_2018} we give separate metrics for these classes.

Tab.~\ref{tab:apollo_per_class} shows that our adaptive downsampling based on semantic boundaries improves overall quality of semantic segmentation. Our approach achieves a mIoU gain of 3-5\% for target classes and up to 2\% overall. This improvement comes at negligible computational cost. Fig.~\ref{fig:apollo cost} shows that our approach consistently produces better results even under fixed computational budgets.

It is not surprising that target classes benefit more. Indeed, focusing on boundaries of some classes may lower performance on other classes. This gives one a flexibility of reflecting importance of certain classes over the others depending on the application. 

\paragraph{CityScapes \cite{Cordts2016Cityscapes}} is another commonly used open road scene dataset providing 5K annotated images of size $1024\times2048$ with 19 classes in evaluation. Following the same test protocol, we evaluated our approach using PSP-Net~\cite{zhao2017pyramid,oandrienko2018fast-semantic-segmentation} (with ResNet50~\cite{he2016deep} backbone) and Deeplabv3+~\cite{chen2018encoder} (with Xception65~\cite{chollet2017xception} backbone) as the base segmentation model. The {mIoU} results are shown in Tab.~\ref{tab:CityScapes results} and Fig.~\ref{fig:cityscapes cost} where we again see consistent improvements of up to 4\%.

\paragraph{Synthia}\label{sec:exp:synthia}\cite{Synthia_2016} is a synthetic dataset of 13K HD images taken from an array of cameras moving randomly through a city. 
The results in Tab.~\ref{tab:synthia results} show that our approach improves upon the baseline model. The cost-performance analysis in Fig.~\ref{fig:synthia cost} shows that our method improves segmentation quality of target classes by 1.5\% to 3\% at negligible cost.

\paragraph{Person segmentation}\label{sec:exp:person}
The Supervisely Person Dataset~\cite{supervisely} is a collection of $5711$ high-resolution images with 6884 high-quality annotated person instances. The dataset set contains pictures of people taken in different conditions, including portraits, land- and cityscapes. We have randomly split the dataset into training ($5140$) and testing subsets ($571$). The dataset has only two labels: person and background. Segmentation results for this dataset are shown in Tab.~\ref{tab:supervisely} with a cost-performance analysis with respect to the baseline shown in Fig.~\ref{fig:supervisely cost}. The experiment shows absolute mIoU increases up to $5.8\%$, confirming the advantages of non-uniform downsampling for person segmentation tasks as well.

\begin{figure*}[h]
    \centering
    \vspace{-1cm}
    \begin{tabular}{c}
    \includegraphics[width=0.85\linewidth,trim=0 82mm 0 4cm,clip] {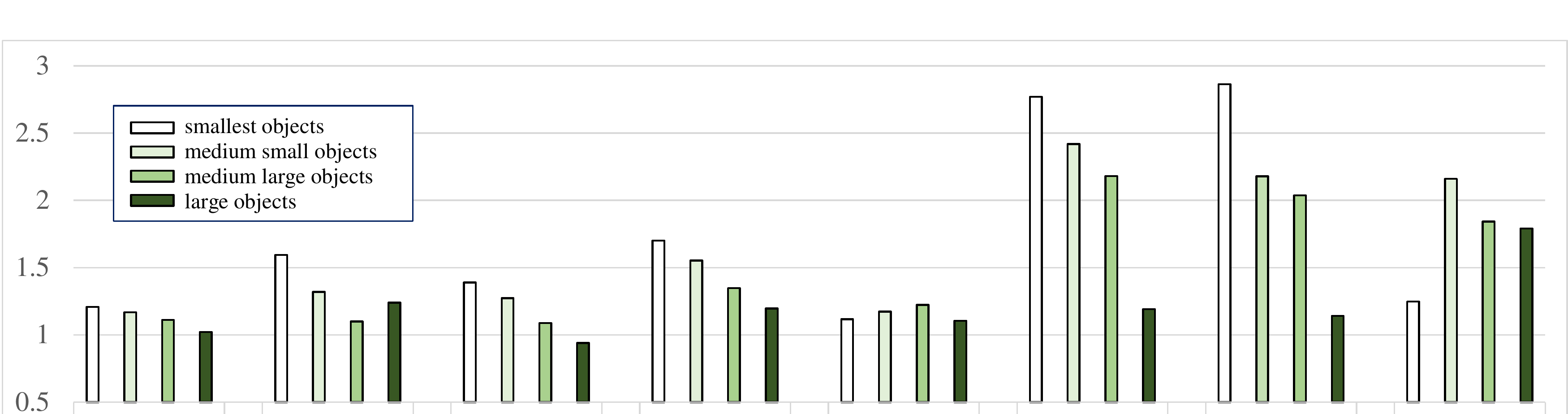}\\[-1ex]
    \includegraphics[width=0.85\linewidth,trim=0 45mm 0 145mm,clip] {object_size_per_class_apollo_3.pdf}
    \end{tabular}
    \vspace{-2ex}
    \caption{Average recall of objects broken down by object classes and sizes on the validation set of ApolloScapes. Values are expressed relative to the baseline. All objects of a class were split into 4 equally sized bins based on objects' area. Smaller bin number correspond to objects of smaller size. The total number of objects in each class is marked by ``\#''. As well as in Fig.~\ref{fig:object_size_apollo} there is negative correlation between object sizes and relative recall for all classes except rare ``rider'' and ``tricycle''.}
    \label{fig:object_size_apollo_per_class}
\end{figure*}

\begin{figure}[t]
    \centering
    \includegraphics[width=0.85\linewidth]{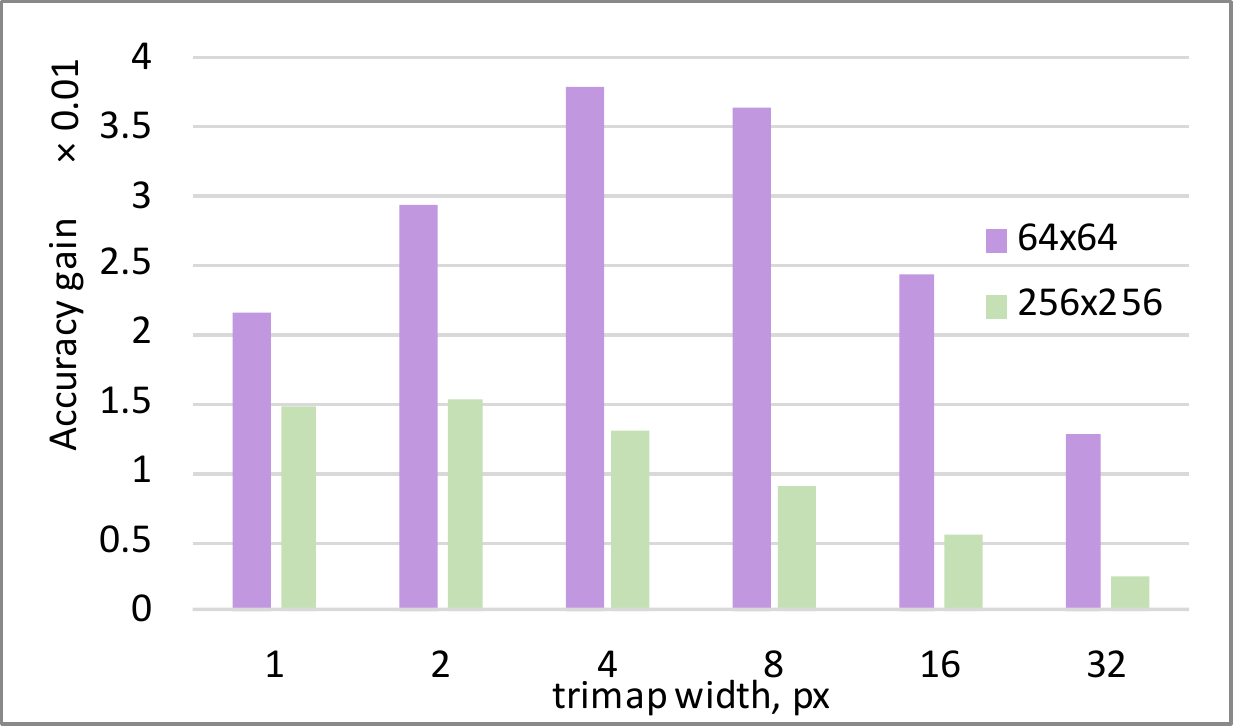}
    \caption{Absolute accuracy difference between our approach and the baseline around semantic boundaries on Supervisely dataset for downsampling resolutions $64 \times 64$ and $256 \times 256$.}
    \label{fig:supervisely trimap 256}
    \label{fig:supervisely trimap 64}
    \label{fig:supervisely trimap}
\end{figure}

\begin{table}[bt]
    \setlength\tabcolsep{4pt}
    \footnotesize
    \centering
    \begin{tabular}{l|cccc|cccc}
    & \begin{sideways}\textbf{\parbox{15mm}{\centering downsample\\[-0.8ex] resolution}}\end{sideways} 
    & \begin{sideways}\parbox{12ex}{\centering\bf\linespread{0}\selectfont auxiliary net\\[-0.8ex] resolution} \end{sideways}  
    & \begin{sideways}\textbf{flops, $\cdot 10^9$}\end{sideways}  
    & \begin{sideways}\textbf{mIoU}\end{sideways}
    & \begin{sideways}\textbf{\parbox{15mm}{\centering downsample\\[-0.8ex] resolution}}\end{sideways} 
    & \begin{sideways}\parbox{12ex}{\centering\bf\linespread{0}\selectfont auxiliary net\\[-0.8ex] resolution} \end{sideways}  
    & \begin{sideways}\textbf{flops, $\cdot 10^9$}\end{sideways}  
    & \begin{sideways}\textbf{mIoU}\end{sideways}  \\
    \hline
    \textbf{backbone}
    & \multicolumn{4}{c|}{PSP-Net\cite{zhao2017pyramid}} 
    & \multicolumn{4}{c}{Deeplabv3+\cite{chen2018encoder}} \\
    \hline
    \textbf{ours} & \multirow{2}{*}{64}  & 32  & 4.37  & \textbf{0.32} 
                  & \multirow{2}{*}{160}  & 32 & 17.54 & \textbf{0.58} \\
    \textbf{baseline} & & -   & 4.20  & 0.29 
                      & & -   & 17.23 & 0.54 \\
    \hline
    \textbf{ours} & \multirow{2}{*}{128} & 32  & 11.25  & \textbf{0.43} 
                  & \multirow{2}{*}{192} & 32  & 25.12  & \textbf{0.62}\\
    \textbf{baseline} & & -  & 11.08  & 0.40
                      & & -  & 24.81  & 0.61\\
    \hline
    \textbf{ours} & \multirow{2}{*}{256} & 32  & 44.22 & \textbf{0.54}
                  & \multirow{2}{*}{224} & 32  & 34.08 & \textbf{0.65}\\
    \textbf{baseline} & & -  & 44.05 & {0.54}  
                      & & -  & 33.77 & {0.62} \\
    \end{tabular}%
  \caption{CityScapes results with different backbones.}
  
  \label{tab:CityScapes results}%
  \vspace{5ex}

    \footnotesize
    \centering
    \begin{tabular}{l|c|c|cc}
     &
     \parbox{11ex}{\centering\textbf{downsample\\[-0.8ex]resolution}}&
     \parbox{6ex}{\centering\textbf{flops,\\[-0.6ex]$\cdot 10^9$}} &
     \parbox{7ex}{\centering\textbf{all\\[-0.8ex]classes}} &
     \parbox{7ex}{\centering\textbf{target\\[-0.8ex]classes}} \\[0.7ex]
    \hline
    \textbf{ours} & \multirow{2}{*}{32}    & 0.38  & \textbf{0.67} & \textbf{0.61} \\
    \textbf{baseline} &   & 0.31  & 0.65  & 0.58 \\
    \hline
    \textbf{ours} & \multirow{2}{*}{64}    & 1.40  & \textbf{0.77} & \textbf{0.73} \\
    \textbf{baseline} & & 1.23  & 0.76  & 0.71 \\
    \hline
    \textbf{ours} & \multirow{2}{*}{128}   & 5.49  & \textbf{0.86} & \textbf{0.83} \\
    \textbf{baseline} & & 4.93  & 0.84  & 0.81 \\
    \hline
    \textbf{ours} & \multirow{2}{*}{256} & 21.85 & \textbf{0.92} & \textbf{0.91} \\
    \textbf{baseline} & & 19.74 & 0.91  & 0.89 \\
    \end{tabular}%
  \caption{Synthia results (mIoU). With the same input resolution our approach improves the segmentation quality.}
  \label{tab:synthia results}%
  \vspace{5ex}

  \footnotesize
  \centering
    \begin{tabular}{l|c|cc|cc}
     &
     \parbox{11ex}{\centering\textbf{downsample\\[-0.8ex]resolution}}&
     \parbox{4ex}{\centering\textbf{flops,\\[-0.6ex]$\cdot 10^9$}} &
     \textbf{mIoU} &
     \parbox{7ex}{\centering\textbf{back-\\[-0.8ex]ground}} &
     \parbox{7ex}{\centering\textbf{person}} \\[0.7ex]
    \hline
    \textbf{ours} & \multirow{2}{*}{16}    & 0.15  & \textbf{0.73} & \textbf{0.84} & \textbf{0.62} \\
    \textbf{baseline} &   & 0.07  & 0.69  & 0.81  & 0.56 \\
    \hline
    \textbf{ours} & \multirow{2}{*}{32} & 0.35  & \textbf{0.76} & \textbf{0.86} & \textbf{0.67} \\
    \textbf{baseline} & & 0.30  & 0.76  & 0.85  & 0.66 \\
    \hline
    \textbf{ours} & \multirow{2}{*}{64} & 1.39  & \textbf{0.83} & \textbf{0.90} & \textbf{0.76} \\
    \textbf{baseline} & & 1.22  & 0.80  & 0.88  & 0.71 \\
    \hline
    \textbf{ours} & \multirow{2}{*}{128} & 5.42  & \textbf{0.87} & \textbf{0.93} & \textbf{0.82} \\
    \textbf{baseline} & & 4.90  & 0.85  & 0.91  & 0.79 \\
    \hline
    \textbf{ours} & \multirow{2}{*}{256} & 20.11 & \textbf{0.90} & \textbf{0.94} & \textbf{0.86} \\
    \textbf{baseline} & & 19.59 & 0.89  & 0.93  & 0.84
    \end{tabular}%
  \caption{Supervisely results. With the same input resolution our approach improves the segmentation quality.}
  \label{tab:supervisely}%
  \vspace{-7ex}
\end{table}%

\subsection{Boundary Accuracy} 
We design an experiment to show that our method improves boundary precision. We adopt a standard trimap approach~\cite{kohli2009robust} where we compute the classification accuracy within a band (called trimap) of varying width around boundaries of segments. We compute the trimap plots for two input resolutions in Fig.~\ref{fig:supervisely trimap 64} for person segmentation dataset described above. Our methods improves mostly in the vicinity of semantic boundaries. Interestingly, for the input resolution of $64\times64$ the maximum accuracy improvement is reached around trimap width of 4 pixels. This may be attributed to the fact that  downsampling model in Sec.~\ref{sec:nus:sampling_model} does not depend on downsampling resolution and essentially defines the same sampling tensor for all sizes of downsampled image. Thus, the distances between neighboring points for $64\times64$ sampling locations are approximately 4 times larger than the respective distances for $256\times256$ sampling locations. This leads to reduced gain of accuracy within narrow trimaps.

\subsection{Effect of Object Size}
\begin{figure}
    \centering
    \includegraphics[width=0.85\linewidth]{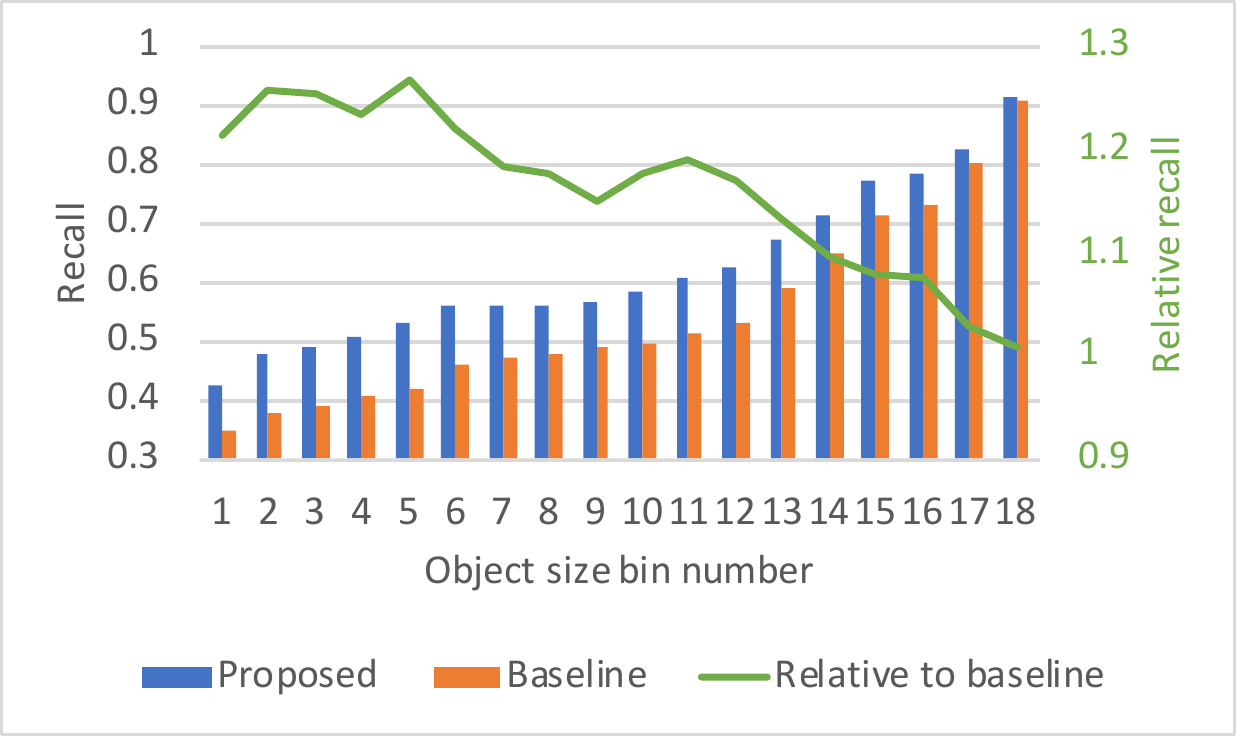}
    \caption{Average recall of objects of different sizes. All objects in the validation set of ApolloScapes were grouped into several equally sized bins by their area. A smaller bin number corresponds to smaller objects. Downsample resolution is $64\!\times\!64$. We  improve baseline more on smaller objects. The green curve (right vertical axis) shows that the relative recall (the average recall of baseline is taken for $1$) is negatively correlated with the object sizes.}
    \label{fig:object_size_apollo}
\end{figure}

Since our adaptive downsampling is trained to select more points around semantic boundaries, it implicitly provides larger support for small objects. This results in better performance of the overall system on these objects. Instance level annotations allow us to verify this by analyzing quality statistics with respect to individual objects. This is in contrast to usual pixel-centric  segmentation metrics (mIoU or accuracy). \Eg, the \emph{recall} of a segmentation of an object is defined as ratio of pixels classified correctly (pixel predicted to belong to the true object class) to the total number of pixels in the object\footnote{Recall usually comes together with \emph{precision}. Since segmentation does not have instance labels, the object-level precision is undefined.}. Fig.~\ref{fig:object_size_apollo_per_class} and \ref{fig:object_size_apollo} show the improvement of recall over baseline for objects of different sizes and categories. Our method degrades more gracefully than the uniform downsampling as the object size decreases.

\section*{Conclusions}

In this work, we described a novel method to perform non-uniform content-aware downsampling as an alternative method to uniform downsampling to reduce the computational cost for semantic segmentation systems. The adaptive downsampling parameters are computed by an auxiliary CNN that learns from a non-uniform sample geometric model driven by semantic boundaries. Although the auxiliary network requires additional computations, the experimental results show that the network improves segmentation performance while keeping the added cost low, providing a better cost-performance balance. Our method significantly improves performance on small objects and produces more precise boundaries. In addition, any off-the-shelf segmentation system can benefit from our approach as it is implemented as an additional block enclosing the system. 

A potential future research direction is employing more advanced interpolation methods, similar to \cite{Weickert2016compression}, which can further improve quality of the final result.

Finally, we note that our adaptive sampling may benefit other applications with pixel-level predictions where boundary accuracy is important and downsampling is used to reduce computational cost. 
This is left for future work.

\appendix

\renewcommand*{\thesection}{Appendix~\Alph{section}}
\section{Non-uniform Sampling Error}\label{sec:supp}

As stated in the submission the error bound decreases as $\mathcal O(\frac{\kappa l^2}{N^2})$ assuming $N$ sampling points are uniformly distributed \emph{near the segment boundary} where $\kappa$ and $l$ are the maximal curvature and length of the boundary respectively. 
To show the upper bound on the best approximation it suffices to show an example that provides $\mathcal O(\frac{\kappa l^2}{N^2})$ boundary approximation error.

\begin{figure}[t]
    \centering
    \includegraphics[width=0.9\linewidth, clip, trim=5cm 7cm 5cm 0, page=2]{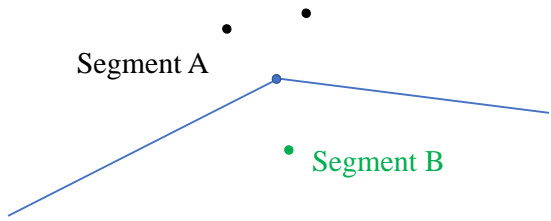}
    \caption{Three sampling points per corner is enough to build any piece-wise linear boundary.}
    \label{fig:three}
\end{figure}

Assuming commonly used linear interpolation method (which we use in the paper) the boundary of segments is piece-wise linear. Using $N$ sampling points we can define any piece-wise linear curve with $M=N/3$ segments, see Fig.~\ref{fig:three}. 

\begin{figure}[t]
    \centering
    \includegraphics[width=\linewidth, clip, trim=0 5cm 0 0]{{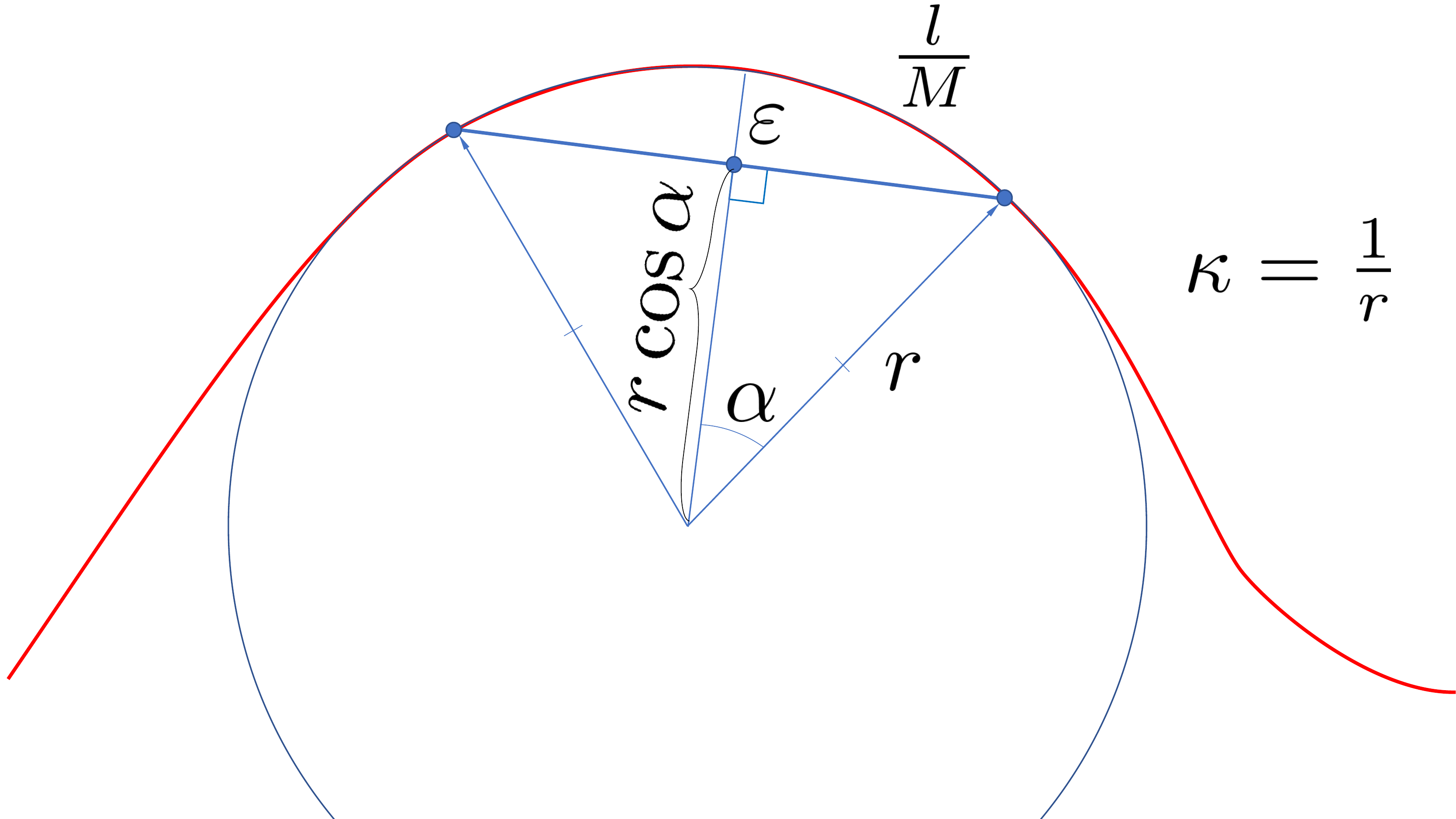}}
    \caption{Illustration for the piece-wise linear approximation for a curve (red) where the ends of linear segments (blue) lie on the curve.}
    \label{fig:approx}
\end{figure}

Let $f(s)$ be a curve of length $l$ in $\mathbb R^2$ for $s_0\le s \le s_1$ and $p(t)$ be its linear approximation with $M$ segments. We define boundary approximation error $\epsilon$ as the maximal distance between the curve and its approximation:
$$\epsilon=\sup_t \inf_s \|p(t) - f(s)\|.$$

We place the ends of the segments of $p$ exactly on curve $f$ such that they are uniformly distributed. That is, each segment encloses a piece of the curve of length $l/M$. The error on each segments can be bounded by approximation error of an arc of radius $r = \frac 1 \kappa$ as shown in Fig.~\ref{fig:approx}:
$$\varepsilon \le \frac{1 - \cos \alpha}{\kappa} = \mathcal O \left(\frac{\kappa l^2}{M^2} \right)$$
where we used the facts
$$\alpha \cdot r= \frac{l}{2M} \;\; \text{and} \;\; \cos\alpha = 1 - \frac{\alpha^2}2  + \mathcal O(\alpha^4). $$
This immediately leads to the bound $\epsilon=\mathcal O(\frac{\kappa l^2}{N^2})$.

{\small
\bibliographystyle{ieee}
\bibliography{egbib}
}

\end{document}